%% file: main.tex
\documentclass[letterpaper, 10 pt, journal, twoside]{IEEEtran}
\input{preamble.tex}

\begin{document}
\title{Voxgraph: Globally Consistent, Volumetric Mapping using Signed Distance Function Submaps}

\author{Victor~Reijgwart*,
        Alexander~Millane*,
        Helen~Oleynikova,
        Roland~Siegwart, Cesar~Cadena,
        Juan~Nieto
\thanks{Manuscript received: September, 10, 2019; Revised November, 12, 2019; Accepted October, 28, 2019.}
\thanks{This paper was recommended for publication by Editor S. Behnke upon evaluation of the Associate Editor and Reviewers' comments.
This work was supported by the National Center of Competence in Research (NCCR) Robotics through the Swiss National Science Foundation, as well as the Defense Advanced Research
Projects Agency (DARPA) under Agreement No. HR00111820045, and Armasuisse.}
\thanks{* The authors contributed equally.}%
\thanks{All authors are with the Autonomous Systems Lab, ETH Z\"urich, Switzerland
        {\tt\footnotesize victorr@ethz.ch; millanea@ethz.ch;
        helenol@ethz.ch;
        rsiegwart@ethz.ch;
        cesarc@ethz.ch;
        nietoj@ethz.ch}}%
\thanks{Digital Object Identifier (DOI): \href{https://doi.org/10.1109/LRA.2019.2953859}{10.1109/LRA.2019.2953859}}
}


\begin{acronym}
\acro{SLAM}{Simultaneous Localization And Mapping}
\acro{SDF}{Signed Distance Function}
\acro{TSDF}{Truncated Signed Distance Function}
\acro{BoW}{Bag of Words}
\acro{MAV}{Micro Aerial Vehicle}
\acro{ESDF}{Euclidean Signed Distance Function}
\acro{AABB}{Axis Aligned Bounding Box}
\acrodefplural{AABB}{Axis Aligned Bounding Boxes}
\acro{OBB}{Oriented Bounding Box}
\acro{ICP}{Iterative Closest Point}
\acro{SGD}{Stochastic Gradient Descent}
\acro{ATE}{Absolute Trajectory Error}
\acro{RMSE}{Root Mean Squared Error}
\acro{IMU}{Inertial Measurement Unit}
\acro{DOF}{Degrees of Freedom}
\end{acronym}

\maketitle

\input{sections/abstract.tex}

\IEEEpeerreviewmaketitle

\markboth{IEEE Robotics and Automation Letters. Preprint Version. Accepted October, 2019}
{Reijgwart \MakeLowercase{\textit{et al.}}: Globally Consistent, Volumetric Mapping using Signed Distance Function Submaps} 

\input{sections/introduction.tex}
\input{sections/related_work.tex}
\input{sections/problem_statement.tex}
\input{sections/system.tex}
\input{sections/frontend.tex}

\input{sections/backend.tex}
\input{sections/experiments.tex}
\input{sections/conclusion.tex}


\bibliographystyle{IEEEtran}
\bibliography{bibliography}

\end{document}

%% file: preamble.tex
\usepackage[nolist,nohyperlinks]{acronym}
\usepackage[utf8]{inputenc}
\usepackage{graphicx}
\usepackage{algorithm,algpseudocode}
\algrenewcommand\algorithmicrequire{\textbf{Input:}}
\algrenewcommand\algorithmicensure{\textbf{Output:}}

\makeatletter
\newcommand\fs@betterruled{%
  \def\@fs@cfont{\bfseries}\let\@fs@capt\floatc@ruled
  \def\@fs@pre{\vspace*{5pt}\hrule height.8pt depth0pt \kern2pt}%
  \def\@fs@post{\kern2pt\hrule\relax}%
  \def\@fs@mid{\kern2pt\hrule\kern2pt}%
  \let\@fs@iftopcapt\iftrue}
\floatstyle{betterruled}
\restylefloat{algorithm}
\makeatother

\usepackage{xcolor}

\usepackage{amsmath}
\usepackage{amsfonts}

\usepackage{bm}
\renewcommand{\vec}[1]{\bm{#1}}

\DeclareMathOperator*{\argmin}{arg\,min}

\newcommand{\refeq}[1]{(\ref{#1})}

\usepackage{booktabs}
\newcommand{\ra}[1]{\renewcommand{\arraystretch}{#1}}
\usepackage{caption}

\usepackage{url}

\usepackage{siunitx}

\usepackage{hyperref}
\usepackage{cleveref}

\usepackage{subcaption}

\usepackage[font=footnotesize,labelfont=bf]{caption}


%% file: sections/abstract.tex
\begin{abstract}
Globally consistent dense maps are a key requirement for long-term robot navigation in complex environments. While previous works have addressed the challenges of dense mapping and global consistency, most require more computational resources than may be available on-board small robots. We propose a framework that creates globally consistent volumetric maps on a CPU and is lightweight enough to run on computationally constrained platforms.
Our approach represents the environment as a collection of overlapping \ac{SDF} submaps, and maintains global consistency by computing an optimal alignment of the submap collection. By exploiting the underlying \ac{SDF} representation, we generate correspondence-free constraints between submap pairs that are computationally efficient enough to optimize the \textit{global} problem each time a new submap is added. We deploy the proposed system on a hexacopter \ac{MAV} with an Intel i7-8650U CPU in two realistic scenarios: mapping a large-scale area using a 3D LiDAR, and mapping an industrial space using an RGB-D camera. In the large-scale outdoor experiments, the system optimizes a 120x80\,m map in less than 4\,s and produces absolute trajectory \acsp{RMSE} of less than 1\,m over 400\,m trajectories. Our complete system, called \textit{voxgraph}, is available as open source\footnote{\url{https://github.com/ethz-asl/voxgraph}}.

\begin{IEEEkeywords}
Mapping,
SLAM,
Aerial Systems: Perception and Autonomy
\end{IEEEkeywords}




\end{abstract}

%% file: sections/introduction.tex
\section{Introduction}
\label{sec:introduction}

\IEEEPARstart{I}{n} order to navigate and interact with their environment, robotic platforms typically build an internal representation of the observed world. This process, \ac{SLAM}, has been a focus of robotics research over past decades~\cite{cadena2016past}. Many successful systems convert input data to a set of features, and use these as the map representation. Recent systems have shown estimation of globally consistent, feature-based maps in real-time~\cite{mur2015orb, schneider2018maplab}. While feature-based maps have proven indispensable for motion estimation on robotic platforms~\cite{burri2015real, qin2018vins}, they are of limited use for tasks beyond localization due to the difficulties of extracting the shape and connectivity of surfaces from a sparse set of samples.

\begin{figure}
    \centering
    \includegraphics[width=0.475\textwidth]{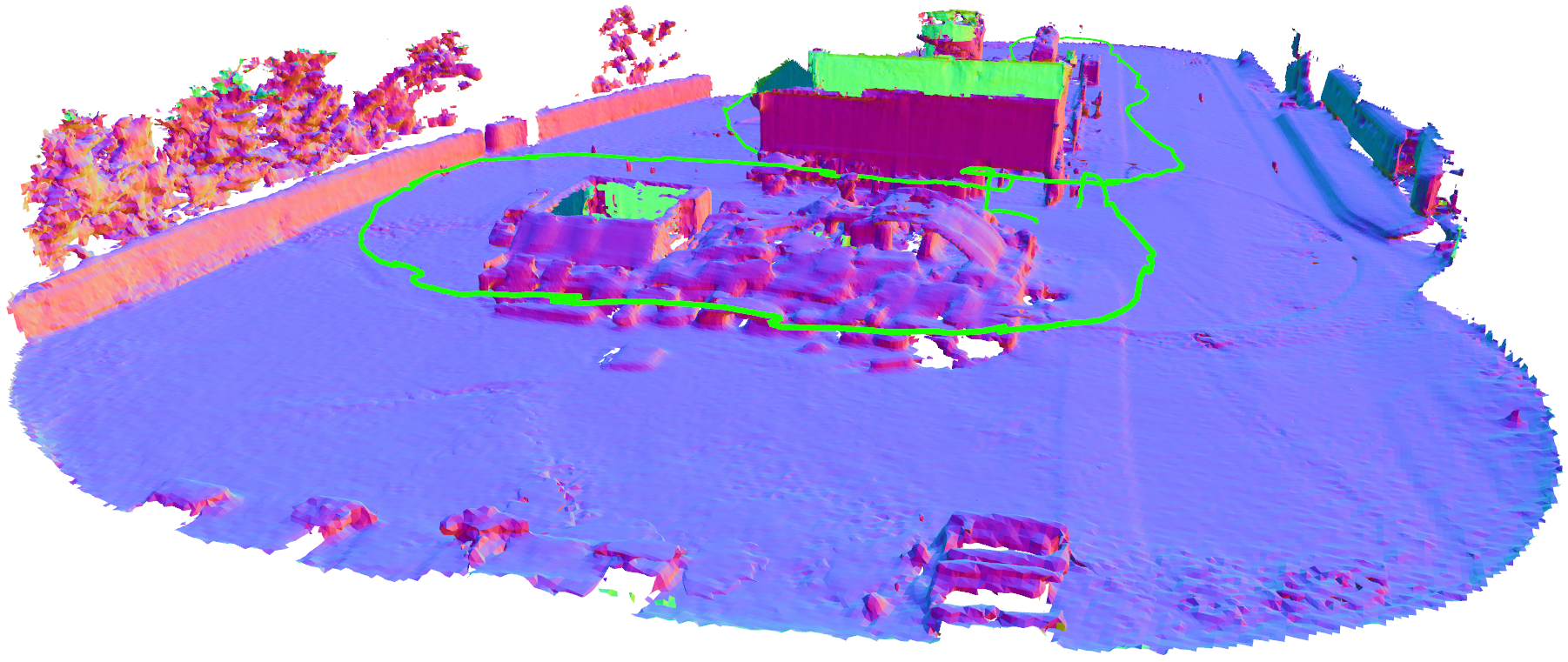}
    \caption{A reconstruction resulting from a  400\,\si{\meter}-long \ac{MAV} flight through the search and rescue training site discussed in Sec.~\ref{sec:experiments_field}. In the foreground a large pile of rubble from a collapsed structure is visible. The trajectory (green) also contains indoor-outdoor transitions through a building (visible behind rubble).}
    \label{fig:arche_reconstruction}
    \vspace{-4mm}
\end{figure}

Several systems build more geometrically complete representations of the environment, producing maps that are useful for tasks such as obstacle avoidance and path planning. Implicit surfaces represented as \acfp{SDF} have emerged as an effective representation for this purpose. The move away from sparse features, however, makes producing a globally consistent map challenging, since global optimization of the map quickly becomes intractable as the amount of data increases. Many existing approaches, therefore, are limited to operation in small-scale environments where drift is limited, or store raw input data such that a globally accurate map can be calculated from scratch as further data becomes available. 

This paper introduces a novel system for building globally-consistent volumetric maps on a CPU. Central to our approach is the representation of the world as a set of overlapping \ac{SDF} submaps. In contrast to similar proposals, we do not compute the full sensor trajectory, but propose a map-centric approach instead. In particular, the proposed system estimates globally consistent submap poses based on the dense map itself, utilizing the underlying \ac{SDF} representation to perform geometric alignment. We propose geometric constraints that are efficient enough to perform global optimization of the map frequently, and in real-time. Furthermore, we formulate the problem as a graph optimization and include odometry and loop closure constraints. In several experiments we show the efficacy of the resulting system for globally consistent large scale dense mapping and demonstrate its use on several \ac{MAV} systems.

In summary the contributions of this paper are:
\begin{itemize}
    \item A novel submap registration method based on the \ac{ESDF}, points on the level-set, and constraint subsampling.
    \item Inclusion of global loop-closures from traditional methods alongside the \ac{SDF} registration constraints, to maintain consistency in case of wide-baseline place revisiting.
    \item Application of an \ac{SDF} submap-based mapping system to dynamic \ac{MAV} flight in a large-scale environment.
    \item Release of an open-source implementation, as well as the challenging, novel dataset used for evaluation: a hexacopter equipped with visual, inertial, LiDAR and RTK GPS sensors flying through a artificial disaster site.
\end{itemize}



%% file: sections/related_work.tex
\section{Related Work}
\label{sec:related_work}

    \ac{SLAM} has received considerable research attention in past decades leading to many successful approaches~\cite{cadena2016past}. While sparse, feature-based representations have been widely adopted, denser representations of geometry are typically required to enable motion planning on robotic systems.

Dense \ac{SLAM} systems have employed a variety of map representations, the choice of which determines many properties of the resulting system. Several successful approaches propose to represent the map as a collection of well-localized camera frames and corresponding depth maps~\cite{engel2014lsd}. 
In contrast, Whelan et. al.~\cite{whelan2012kintinuous} suggests attaching a deformable mesh to a pose-graph representing the past trajectory of a camera. The same authors, in more recent work~\cite{whelan2015elasticfusion}, dispose of trajectory estimation altogether and represent the map as a deformable collection of surfels. These systems provide a richer understanding of the occupied space, yet they do not explicitly  distinguish  free  space  from  unknown  space,  which limits their applicability for autonomous navigation. Conversely, volumetric maps store information about both free and occupied space. Occupancy grid mapping, in particular its efficient implementation in 3D~\cite{hornung2013octomap}, has seen widespread adoption on robotic platforms~\cite{burri2015real}.

An alternative volumetric representation is based on implicit representations of surface geometry through \acp{SDF}, introduced in~\cite{curless1996volumetric}. The efficacy of these representations in fusing high-rate, noisy depth data from consumer-grade depth cameras, shown first by Newcombe et. al~\cite{izadi2011kinectfusion}, has led to a rapid rise in their popularity. Furthermore, the advantages of this representation have led to their recent adoption on robotic platforms~\cite{lin2018autonomous, oleynikova2017voxblox, wagner2014graph}, 
where the distance field has shown additional utility for optimization-based motion planning~\cite{oleynikova2018safe, ratliff2009chomp}. However, in order to perform fusion of high-rate data into a volumetric representation, most approaches discard raw observational data. This information loss is problematic for maintenance of a globally-consistent map. In particular, the lack of a systematic approach for correcting the map to reflect new information relating past poses, loop-closure being a prototypical example, means that \ac{SDF} maps are susceptible to corruption at a global scale.

Researchers have suggested a number of approaches to mitigate this problem. BundleFusion~\cite{dai2017bundlefusion} proposes to maintain a globally-consistent \ac{TSDF}-based reconstruction by storing raw input data, globally optimizing the sensor trajectory with each arriving frame, and frequently re-integrating stored sensor data into the global \ac{SDF}. This approach produces state-of-the-art results in terms of reconstruction quality, but requires significant computational resources. Several recent works~\cite{maier2017efficient, han2018flashfusion} have proposed efficiency improvements to the original scheme, principally by utilizing keyframes and computing the global reconstruction as a function of these keyframes only, as well as speed-ups to the original global optimization scheme. In contrast to our proposal however, these works suggest to maintain a single monolithic map.

Another approach is to represent the reconstructed environment as a collection of submaps. Corrections to the global map are performed by adjusting the relative submap positions. This approach avoids storage and re-integration of raw sensor data. One distinguishing factor among submap-based systems is the policy for submap generation. Several works~\cite{henry2013patch, kahler2016real} suggest \emph{partitioning} the environment into submaps with the goal of minimizing subvolume overlap. Sensor tracking proceeds by performing frame-to-model alignment on all submaps within the sensor viewing frustum. Global consistency is maintained by optimizing a graph linking submaps and past poses through observation data. The advantage of spatial \emph{partitioning} is that the size of the map remains bounded given a scene of bounded size. However, sensor tracking in several submaps simultaneously remains an expensive operation, particularly on systems lacking a GPU.

Another approach is to generate submaps continuously, resulting in a steady stream of new submaps, each potentially overlapping with existing submaps to an arbitrary degree. The advantage of this approach is that the sensor pose, at the time of frame integration, need only to be tracked with respect to the current submap. Several works follow this approach~\cite{choi2015robust, wagner2014graph, hess2016real} employing variations on the frame-to-submap based graph optimization scheme to maintain global consistency of the resulting submap collection. In a different approach, Fioraio et. al.~\cite{fioraio2015large} propose a map-centric system in which the past trajectory is not considered during optimization of the dense map. Instead, submaps are constrained directly to one another through geometric alignment. We extend this work by proposing to use correspondence-free alignment based on the \ac{ESDF}, and propose a weighted subsampling scheme to achieve real-time performance on computationally limited platforms. Furthermore, we include loop closure constraints to maintain consistency in the presence of wide baseline loops and show the suitability of our approach for \ac{MAV} based mapping, using both RGB-D and LiDAR setups.



%% file: sections/problem_statement.tex
\section{Preliminaries}
\label{sec:preliminaries}

We parameterize the transformation between coordinate frames $A$ and $B$ as $T_{BA} \in \text{SE}(3)$, where $T_{BA}: \mathbb{R}^3 \to \mathbb{R}^3$ maps points in frame $A$ to frame $B$ such that $\vec{p}_B = T_{BA} \vec{p}_A.$ Transformations are concatenated as $T_{AC} = T_{AB} T_{BC}$. We express the logarithmic map as $\log : \text{SE}(3) \to \text{se}(3)$, which maps rotational elements to the corresponding \textit{lie algebra} around the identity element.

We assume that the odometry front-end provides estimates in which the direction of gravity is fully observable. This allows us to align several frames in our framework with the gravity vector. Therefore several transformations live in $\mathbb{R}^3 \times \text{SO}(2)$ rather than $\text{SE}(3)$. We will note where this is the case.


\begin{figure*}[ht]
    \centering
    \includegraphics[width=2\columnwidth]{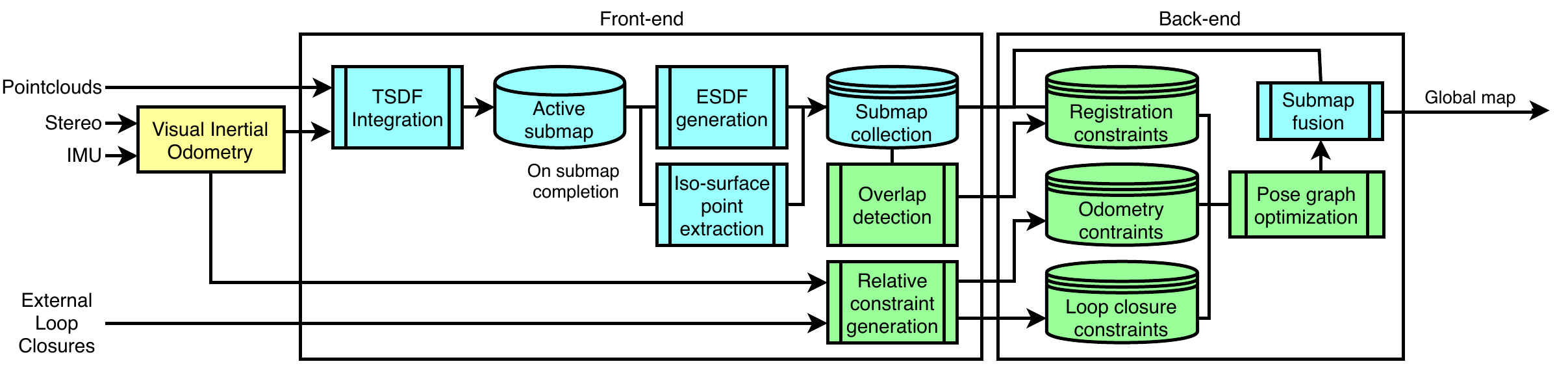}
    \caption{Overview of the proposed system, which generates consistent volumetric maps based on pointclouds, odometry input and optional loop closures. It is split into a front-end and a back-end, which are discussed in Sec.~\ref{sec:front-end} and Sec.~\ref{sec:back-end}, respectively.}
    \label{fig:system_overview_diagram}
    \vspace{-4mm}
\end{figure*}

\section{Problem Statement}
\label{sec:problem_statement}

Given a sequence of pointclouds produced by a sensor (frame $C$) moving through the world (frame $W$, gravity aligned) we aim to build a consistent, \ac{SDF}-based reconstruction. At time step $i$ we have an estimate of the sensor frame's position from odometry, with respect to a local frame $O$, parameterized as a rigid transformation $T_{OC^i} \in \text{SE}(3)$. If $T_{OC^i}$ was precisely known at the time of pointcloud capture, existing techniques~\cite{izadi2011kinectfusion, niessner2013real} could be used to build a monolithic \ac{TSDF} map in $O$. For practical odometry sources, however, $T_{OC^i}$ accumulates error over time. Following reasoning in our previous work~\cite{millane2018cblox}, we build a series of submaps $\{\mathcal{S}_i\}_{i=1}^{N}$ with attached frames $\{S^i\}_{i=1}^N$ from contiguous sequences of sensor data. Submaps are defined by their pose in the world $T_{WS^i} \in \mathbb{R}(3) \times \text{SO}(2)$ and an \ac{SDF}. The \ac{SDF} of a submap $\mathcal{S}_i$ is composed of distance and weight functions $\Phi_{\mathcal{S}_i}: \mathbb{R}^3 \to \mathbb{R}$ and $\omega_{\mathcal{S}_i}: \mathbb{R}^3 \to \mathbb{R}$, which respectively map points in $\mathbb{R}^3$ to $d$, a signed distance to the nearest observed surface, and $w$, a weighting/confidence measure. Under this setup, the reconstruction problem is reduced to determining the optimal submap poses $\{T_{WS^i}\}_{i=0}^N$.




%% file: sections/system.tex
\section{System}
\label{sec:system}

\textit{Voxgraph} is divided into a front-end and a back-end (see Fig.~\ref{fig:system_overview_diagram}). The front-end (Sec.~\ref{sec:front-end}) converts incoming sensor measurements into submaps, as well as constraints to be passed to the back-end. The back-end (Sec.~\ref{sec:front-end}) maintains this set of constraints and estimates the most likely submap collection alignment.

%% file: sections/frontend.tex
\section{Front end}
\label{sec:front-end}

\subsection{Submap Creation}
Contiguous sequences of input pointclouds are combined into submaps by ray casting into a spatially hashed voxel grid (see our previous work~\cite{oleynikova2017voxblox, millane2018cblox} for details). The voxel grid origin is co-located and aligned with its attached frame $S$, which is parameterized by its pose with respect to the world frame $T_{WS}$. We create new submaps at a fixed frequency ($N$ time steps) following from the assumption that errors in odometry estimates accumulate slowly and smoothly, and therefore submaps that are created over sufficiently short timescales remain internally consistent. During construction we also store the sensor trajectory through the submap as a collection of submap-relative poses, 
\begin{equation}
    \mathcal{T}_{\mathcal{S}_i} = \{ T_{S^i C^j},\dots, T_{S^i C^{j+N}}\}.
    \label{eq:pose_history}
\end{equation}
Finally, following submap completion, we compute additional information used by the back-end for submap registration. In particular, the submap's \ac{ESDF} $\Phi_{\mathcal{S}}$ is computed from its \ac{TSDF} as described in \cite{oleynikova2017voxblox}. This process propagates Euclidian distances outside the truncation band used by the TSDF, allowing distance lookups further from surfaces. Furthermore, a collection of isosurface points $\mathcal{U}_{\mathcal{S}}$ are calculated through the marching cubes algorithm~\cite{lorensen_1987_marching-cubes}.

\subsection{Constraint Generation}

The front-end signals the back-end to add constraints to the underlying optimization problem. We implement three kinds of constraints: odometry, registration and loop closure constraints.

\subsubsection{Odometry}
\label{sec:frontend_odometry}

We penalize successive submaps' deviation from their odometry-estimated relative pose. In particular for submaps $\mathcal{S}_i$ and $\mathcal{S}_{i+1}$ joined by the sensor frames $\{C_l\}_{l=k}^{k+N}$, we estimate the submaps' relative transformation through concatenation of odometry estimates as
\begin{equation}
    \hat{T}_{S^i S^{i+1}}
    =
    T_{C^k C^{k+1}} T_{C^{k+1} C^{k+2}} \dots T_{C^{k+N-1} C^{k+N}}.
\end{equation}
This estimate is passed to the back-end to constrain the relevant submap frames (see Sec.~\ref{sec:backend_odometry}).

\subsubsection{Loop Closure}
\label{sec:frontend_loop_closure}
Our system accepts loop closures from external sources. Note that the system is agnostic to the source of loop closures; we tested, for example, DBoW2~\cite{GalvezTRO12} in Sec.~\ref{sec:experiments_rgbd}. Input loop closures take the form of an estimated transformation $\hat{T}_{C^l C^k}$ linking the sensor frame $C$ at time instances $l$ and $k$ (which are not contained within the same submap).

Loop closures relate the sensor position at arbitrary time instances, and therefore generally do not relate submap frames \textit{directly}. To accommodate this we determine the two submaps, $\mathcal{S}_i$ and $\mathcal{S}_j$ which contained these time steps, and look up the submap-relative poses $T_{S^i C^l}$ and $T_{S^j C^k}$ in \eqref{eq:pose_history}. These estimates are passed to the back-end (see Sec.~\ref{sec:backend_loopclosure}).

\subsubsection{Registration}
\label{sec:frontend_registration}

Central to this paper's approach to maintaining consistency is geometric alignment of all overlapping submaps. The front-end detects pairs of overlapping submaps using \acp{AABB} (see \cite{christer_ericson_2005_collision-detection}). In particular an \ac{OBB} in the submap frame $S$ is computed by finding the maximal dimensions of allocated voxel \textit{blocks}~\cite{oleynikova2017voxblox} in each of the axes of $S$. This operation is performed once per submap following pointcloud integration. Then, before global optimization, we compute for each submap an \ac{AABB} which bounds the \ac{OBB} and is aligned with the world frame $W$ using the current estimate of the submap pose $T_{W S^i}$. Registration constraints are created for all overlapping submap pairs, and then sent to the back-end (see Sec.~\ref{sec:backend_registration}).


%% file: sections/backend.tex
\section{Back end}
\label{sec:back-end}

The back-end aligns the submap collection by minimizing the total error of all pose graph constraints. We solve the non-linear least squares minimization
\begin{align}
    \label{eq:optimization}
    \argmin_{\mathcal{X}}
    &
    \sum_{(i,j)\in \mathcal{R}} || e^{i,j}_{\text{reg}}(T_{W S^i}, T_{W S^j}) ||^2_{\sigma_{r}} + \\
    &
    \sum_{(i,j)\in \mathcal{O}} || \vec{e}^{i,j}_{\text{odom}}(T_{W S^i}, T_{W S^j}) ||^2_{\Sigma_{O}} + \nonumber \\
    &
    \sum_{(i,j)\in \mathcal{L}} || \vec{e}^{i,j}_{\text{loop}}(T_{W S^i}, T_{W S^j}) ||^2_{\Sigma_{L}} \nonumber
\end{align}
where 
\begin{equation}
    \mathcal{X}
    =
    \{ T_{WS^1}, T_{WS^2}, \dots, T_{WS^N} \}
\end{equation}
are the submap poses, $T_{W S^i} \in \mathbb{R}^3 \times SO(2)$, and $\mathcal{O}$, $\mathcal{R}$ and $\mathcal{L}$ are sets containing submap index pairs joined by odometry, registration and loop-closure constraints respectively. The squared Mahalanobis distance for the odometry and loop closure constraints is written as $||\vec{e}||^2_{\Sigma_{\mathcal{O}}}$ and $||\vec{e}||^2_{\Sigma_{\mathcal{L}}}$, where $\Sigma_{O}$ and $\Sigma_{L}$ represent their covariance matrices. For the registration constraints $||e||^2_{\sigma_{\mathcal{R}}}$ corresponds to the squared weighted total distance, with scalar weight $\sigma_{\mathcal{R}}$. Note that the optimization variables $T_{W S^i}$ live on a non-linear manifold. We therefore solve for optimal \textit{increments} in the corresponding lie-algebra  (see for example~\cite{dellaert2017factor}). For clarity, however, we treat the optimization as directly over manifold elements in the following sections.

\subsection{Odometry Residuals}
\label{sec:backend_odometry}

Odometry residuals are implemented as cost-terms on the relative pose between submap base-frames, such that for general submaps $\mathcal{S}_i$ and $\mathcal{S}_j$ we have,
\begin{equation}
    \vec{e}^{i,j}_{\text{odom}}(T_{WS^i}, T_{WS^j})
    =
    \log({\hat{T}^{-1}_{S^i S^j} T_{WS^i}^{-1} T_{WS^j}}).
\end{equation}
In the less general case of \textit{odometry} constraints $j=i+1$ and $\hat{T}_{S^i S^{i+1}}$ is the odometry-estimated relative pose calculated in the front-end.

\subsection{Loop Closure Residuals}
\label{sec:backend_loopclosure}

Loop closure residuals constrain the relative pose of points \textit{within} two submaps which are not co-located with submap frames themselves. For $\mathcal{S}_i$ and $\mathcal{S}_j$ and sensor frames $C^l$ and $C^k$ we have,
\begin{align*}
    &\vec{e}^{i,j}_{\text{loop}}(T_{WS^i}, T_{WS^j})
    =\\
    &\qquad\log(\hat{T}_{C^l C^k} (T_{W S^j} T_{S^j C^k})^{-1} T_{W S^i} T_{S^i C^l})
\end{align*}
where $T_{S^i C^l}$ and $T_{S^j C^k}$ are the poses of the sensor at time steps $l$ and $k$ with respect to their submap frames (see Sec.~\ref{sec:frontend_loop_closure}).

\subsection{Registration Constraints}
\label{sec:backend_registration}
In this section we describe our method for generating efficient pairwise registration constraints for submaps produced by the front-end.
We adopt the reasoning behind the \ac{ICP} algorithm (that two submaps can be aligned by minimizing the Euclidean distance between points on their surfaces) and adapt it to utilize the underlying \ac{SDF} representation to perform registration efficiently.

Submaps to be aligned by the back-end have a set of zero-level iso-surface points $\mathcal{U}_{\mathcal{S}}$, as well as an \ac{ESDF} $\Phi_{\mathcal{S}}$, extracted in the front-end (Sec.~\ref{sec:front-end}). One approach to generating pair-wise constraints would be to determine point-to-point correspondences between the iso-surface points of submap pairs, and minimize distance between these pairs. In this commonly used approach, the correspondence search can be the most expensive stage.

\begin{figure}
    \centering
    \includegraphics[width=0.35\textwidth]{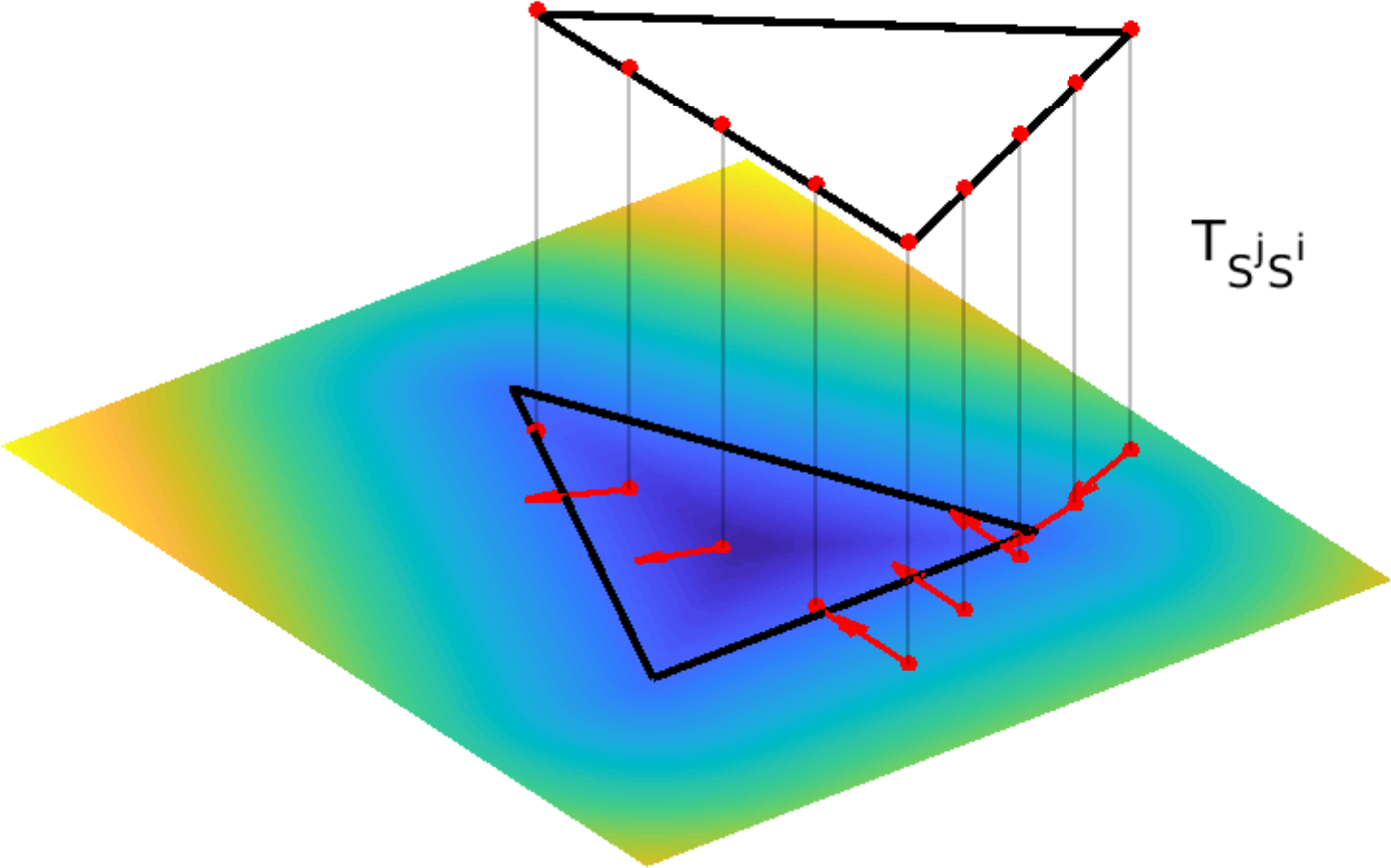}
    \caption{Correspondence-free alignment for a simplified example. The figure shows a two dimension triangle and its corresponding \ac{ESDF} (colored plane). The second triangle is transformed with respect to the original (and offset in z for clarity). Red iso-surface points are transformed into the \ac{ESDF} using $T_{S^jS^i}$ and generate a cost equal to its value. The gradients of the evaluated cost are shown as red arrows.}
    \label{fig:registration_dummy}
    \vspace{-4mm}
\end{figure}

We propose to use correspondence-free alignment. In particular, for a point on the iso-surface of $\mathcal{S}_i$ we are able to determine the distance to the iso-surface of $\mathcal{S}_j$ by reading the \ac{ESDF} value at that point. The registration constraint between two submaps then expresses the total squared distance difference evaluated for all reading points $\vec{p}_{\mathcal{S}_i}^m$ of submap $\mathcal{S}_i$,
$$
e^{i,j}_{\text{reg}}(T_{WS^i}, T_{WS^j})
=
\sum_{m=0}^{N_{\mathcal{S}_i}} r_{{\mathcal{S}_i}{\mathcal{S}_j}}(\vec{p}_{\mathcal{S}_i}^m, T_{{S^j}{S^i}})^2,
$$
where $N_{\mathcal{S}_i}$ is the total number of iso-surface points $\mathcal{U}_{\mathcal{S}_i}$ of submap $\mathcal{S}_i$. The residual $r_{\mathcal{S}_i\mathcal{S}_j}$ is given by
\begin{align*}
    r_{\mathcal{S}_i\mathcal{S}_j}(\vec{p}_{\mathcal{S}_i}^m, T_{{S^j}{S^i}})
    &=
    \Phi_{\mathcal{S}_i}(\vec{p}_{\mathcal{S}_i}^m) - \Phi_{\mathcal{S}_j}(T_{{S^j}{S^i}}\ \vec{p}_{\mathcal{S}_i}^i) \\
    &=
    -\Phi_{\mathcal{S}_j}(T_{{S^j}{S^i}}\ \vec{p}_{\mathcal{S}_i}^m)
\end{align*}
where $T_{{S^j}{S^i}}$ is a function of optimization variables in $\mathcal{X}$ through
\begin{equation}
    T_{{S^j}{S^i}}=T_{W{S^j}}^{-1} T_{W{S^i}}^{\phantom{-1}}. 
\end{equation}
Note that $\Phi_{\mathcal{S}_i}(\vec{p}_{\mathcal{S}_i}^m)=0$ for all points $\vec{p}_{\mathcal{S}_i}^m$, since these points lie on the zero-level surface of ${\mathcal{S}_i}$ by construction.  Figure~\ref{fig:registration_dummy} shows a simplified example of a registration constraint.

In addition to the loss function itself, the solver also uses its first derivative. This Jacobian is obtained by differentiating the residual over the $x,y,z$ and $\textit{yaw}$ coordinates of the pose of the submaps ${\mathcal{S}_i}$ and ${\mathcal{S}_j}$. For brevity, we represent the combination of these coordinates with $\vec{q}$ and state the general form of the solution, namely
\begin{align*}
    \frac{\partial r_{{\mathcal{S}_i}{\mathcal{S}_j}}(\vec{p}_{\mathcal{S}_i}^m)}{\partial \bm{q}}
    &=
    -\frac{\partial \left[\Phi_{\mathcal{S}_j}(T_{{S^j}{S^i}}\ \vec{p}_{\mathcal{S}_i}^m)\ \right]}{\partial \bm{q}}\\
    &=
    - \frac{\partial \Phi_{\mathcal{S}_j}(\vec{s})}{\partial \vec{s}}
    \bigg\rvert_{\vec{s}=T_{{S^j}{S^i}}\ \vec{p}_{\mathcal{S}_i}^m}
    \frac{\partial \left(T_{{S^j}{S^i}}\ \vec{p}_{\mathcal{S}_i}^m\right)}{\partial \bm{q}}.
    \
\end{align*}
The first term corresponds to the derivative of the \ac{ESDF}'s trilinear interpolation function \cite{kang_2006_color-technology}, given by
\begin{equation*}
    \Phi_{\mathcal{S}_j}(\vec{s}) = \vec{g}^T \bm{B}^T \bm{h}(\vec{s})
\end{equation*}
where vector $\vec{g}$ holds the distances of the $8$ voxels surrounding point $\vec{p}$, matrix $\bm{B}$ represents a constant $8$ by $8$ binary matrix and vector function $\bm{h}$ corresponds to
\begin{equation*}
    \vec{h} =
    \left[
        1\;\; 
        \Delta x\;\;
        \Delta y\;\;
        \Delta z\;\; 
        \Delta x \Delta y\;\; 
        \Delta y \Delta z\;\; 
        \Delta z \Delta x\;\; 
        \Delta x \Delta y \Delta z
    \right]^T
\end{equation*}
where $\Delta x$, $\Delta y$ and $\Delta z$ are the relative distances from $\vec{s}$ to its lower left voxel neighbour $\vec{v}$. For example, $\Delta x = (\vec{s}_x-\vec{v}_x)/r$ with $r$ corresponding to the voxel width.


The first term, $\partial \Phi_{\mathcal{S}_j}(\vec{s}) / \partial \vec{s}$, can thus conveniently be computed as a byproduct of the interpolation operations used to get the distance itself. The second term, $\partial \left(T_{{S^j}{S^i}}\ \vec{p}_{\mathcal{S}_i}^m\right) / \partial \vec{q}$, describes how point $\vec{p}_{\mathcal{S}_i}^m$ moves through submap ${\mathcal{S}_j}$ as the relative transformation $T_{{S^j}{S^i}}$ gets updated, and is well known.

\subsection{Registration constraint sub-sampling}
\label{sec:constraint_subsampling}

Given the typically large number of points on iso-surfaces within each submap, joint minimization of the registration error for all surface points of all pairs of overlapping submaps is computationally expensive, even for modestly sized maps. In this section, we describe our proposal for increasing the computation efficiency of pairwise registration constraints through sub-sampling.


We propose to approximate the registration constraints by using only a random subset of their residuals within each solver iteration. The resulting nonlinear optimization shares some similarities with Stochastic Gradient Descent (SGD), 
which has seen successful application to \ac{SLAM} problems in the past~\cite{olson_2006_fast-iterative}. 
For each iso-surface point $\vec{p}_{\mathcal{S}_i}^m$ in $\mathcal{U}_{\mathcal{S}_i}$ we determine a weight $\omega_{\mathcal{S}_i}(\vec{p}_{\mathcal{S}_i}^m)$, by interpolating the weights of the voxels that surround it. The voxel weights provide a confidence measure on the distance estimated by each voxel. In a given solver iteration $k$, the registration error for a submap pair ${\mathcal{S}_i},{\mathcal{S}_j}$ can be approximated by evaluating the registration error over the iso-surface point sub-set $\mathcal{V}_k$. We obtain the subset $\mathcal{V}_k$ by drawing $N_{\mathcal{V}_k}$ samples from $\mathcal{U}_{\mathcal{S}_i}$ with replacement, where the probability of drawing each point $v_{\mathcal{S}_i}^m$ is proportional to its weight $w = \omega_{\mathcal{S}_i}(v_{\mathcal{S}_i}^m)$ (see Algo.~\ref{algo:approximate-registration-cost}).

\begin{algorithm}
\caption{\strut Approximates the registration cost between two submaps using weighted sampling.}
\label{algo:approximate-registration-cost}
\begin{algorithmic}[1]
\Require{\ Submap $\mathcal{S}_i$'s iso-surface point set $\mathcal{U}_{\mathcal{S}_i}$,
\Statex\hspace{\algorithmicindent} their weights $\omega_{\mathcal{S}_i}$ and the number of points $N_{\mathcal{S}_i}$
\Statex\hspace{\algorithmicindent} Submap $\mathcal{S}_j$'s \ac{SDF} $\Phi_{\mathcal{S}_j}$
\Statex\hspace{\algorithmicindent} Sampling ratio $\alpha$
\Statex\hspace{\algorithmicindent} Relative transformation $T_{{S^j}{S^i}}$}
\Ensure{Approximation of $e^{i,j}_{\text{reg}}(T_{WS^i}, T_{WS^j})$}
\Statex
\Function{RegistrationErrorTerm}{}
\State $e^{ij} \gets 0$
\For{$m \gets 1$ to $\alpha N_{\mathcal{S}_i}$}
\State draw $\bm{v}^m$ from $\mathcal{U}_{\mathcal{S}_i}$ with probability $\propto \omega_{\mathcal{S}_i}(\bm{v}^m)$
\State $e^{ij} \gets e^{ij} + \Phi_{\mathcal{S}_j}(T_{{S^j}{S^i}}\ \bm{v}^m)^2$
\EndFor
\State $e^{ij} \gets \frac{1}{\alpha} e^{ij}$
\State \Return{$e^{ij}$}
\EndFunction
\end{algorithmic}
\end{algorithm}


%% file: sections/experiments.tex
\section{Experiments}
\label{sec:experiments}

In this section we aim to validate our claim that the proposed system efficiently generates globally consistent \ac{SDF} maps. To evaluate reconstruction performance we present simulation experiments (Sec.~\ref{sec:experiments_sim}) where access to ground truth geometry is readily available. In Sec.~\ref{sec:experiments_field} we present field experiments conducted on an \ac{MAV} flying through a large outdoor scene and evaluate trajectory accuracy as a proxy for reconstruction performance. Lastly we show qualitative results from an RGB-D equipped \ac{MAV} mapping an industrial space.

\subsection{Simulation Experiments}
\label{sec:experiments_sim}

In this section we present a quantitative analysis of the proposed constraint sub-sampling strategy. We simulated a LiDAR-equipped \ac{MAV} flying a $242~\si{\second}$ trajectory around a multi-story building using Rotors simulator\footnote{\url{https://github.com/ethz-asl/rotors_simulator}}. Drift is simulated by perturbing ground truth odometry with additive, biased Gaussian noise, which produces position and yaw errors which accumulate over time. The final error in the simulated odometry output, used as input to \textit{voxgraph}, is $9.88~\si{\meter}$.

We measure system performance in terms of reconstruction error. In particular, we fuse all submaps into a single map in the Global frame $G$. We then compute a ground truth \ac{SDF} $\Omega_{GT}$ from the simulation environment using an open-source tool developed for this purpose\footnote{\url{https://github.com/ethz-asl/voxblox_ground_truth}}. 
As the ground-plane is a large, geometrically simple object that is well reconstructed by all systems, it tends to dominate the average error, pushing it towards a low constant value. We therefore exclude it from the evaluation to highlight trends more effectively. 
We calculate the \ac{RMSE} error,
\begin{equation}
    \mathcal{E}_p
    =
    \left( \sum_{\bm{p}_G \in \mathcal{O}} \left[ \Phi_{G}(\bm{p}_G) - \Phi_{GT}(\bm{p}_G) \right]^{2} \right)^{1/2},
\end{equation}
where $\Phi_{G}(\bm{p}_G)$ and $\Phi_{GT}(\bm{p}_G)$ are the SDFs of the estimated and ground truth global maps evaluated at $\bm{p}_G$. The set of all observed voxels in fused map is denoted $\mathcal{O}$.

\begin{figure}
    \centering
    \includegraphics[width=0.475\textwidth,trim=0 0 0 28]{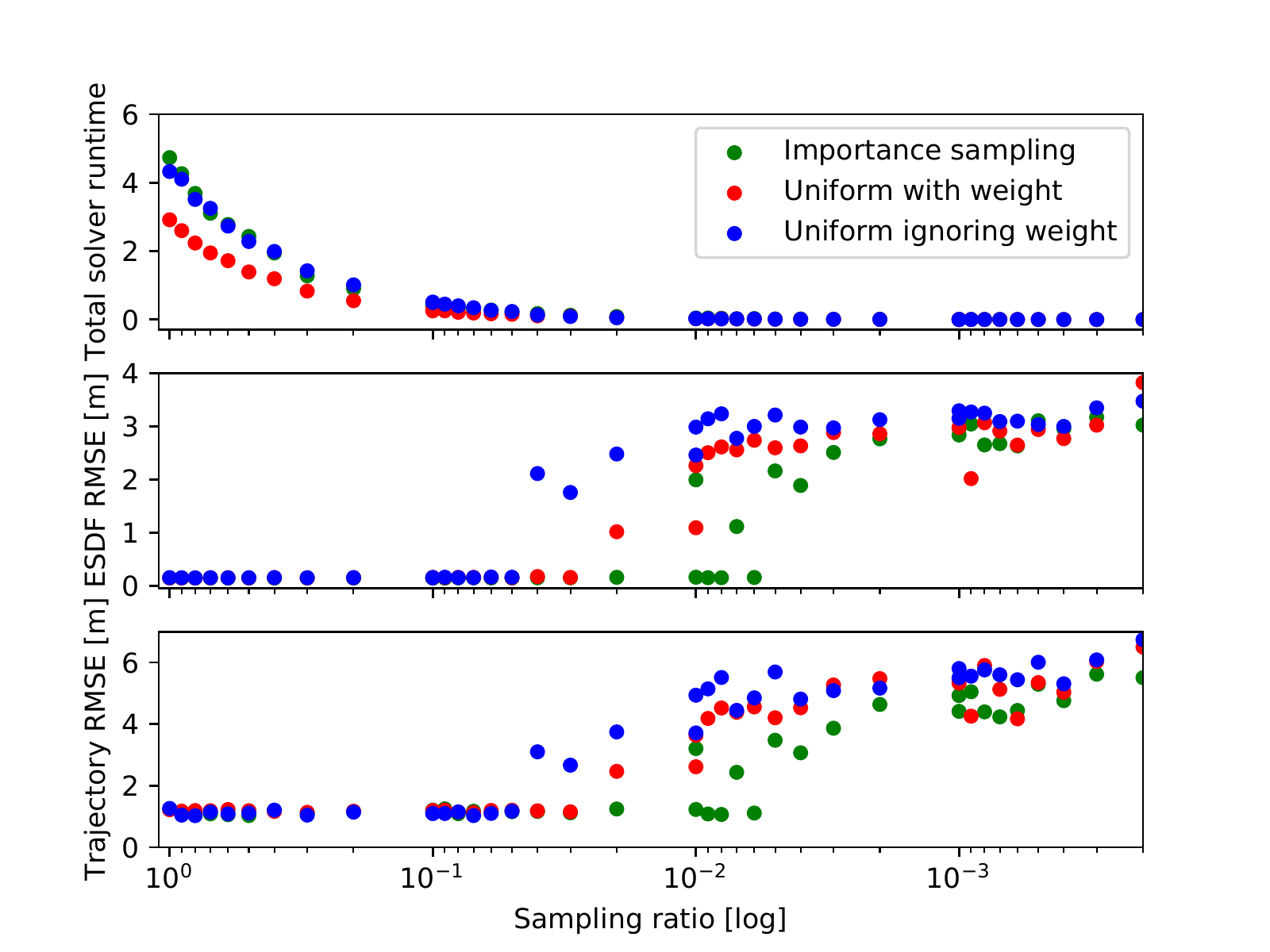}
    \caption{Solver runtime (normalized over the dataset duration), \ac{RMSE} and position errors for various constraint sub-sampling strategies and ratios resulting from a simulation study (Sec.~\ref{sec:experiments_sim}). The figure shows that, in this experiment, there is no observable performance degradation until constraints are sub-sampled at below 5\%.}
    \label{fig:subsampling_reconstruction}
    \vspace{-5mm}
\end{figure}

We perform $120$ experiments in which we vary the sub-sampling ratio between 100\% and 0.01\% and compare the weighted sampling strategy introduced in Sec.~\ref{sec:constraint_subsampling} to two alternative methods. In the first, we uniformly subsample the points $\mathcal{U}_{\mathcal{S}}$ but use their weights in their cost residuals. In the second alternative strategy, we perform uniform subsampling and discard the weights altogether.  \Cref{fig:subsampling_reconstruction} shows the normalized runtime, reconstruction, and position \ac{RMSE} for these experiments. As one might expect, pose graph optimization run-time decreases linearly with the sampling ratio. Between sub-sampling ratios of $100\%$ and ${\sim}5\%$ there is no-observable change in the reconstruction or position errors. This validates that constraint sub-sampling can be performed to increase computational efficiency without incurring significant costs in terms of performance. Furthermore, it shows that weighted sampling yields more reliable results when high downsampling ratios are used. For the remainder of this paper we therefore use weighted sub-sampling and a sampling ratio of $5\%$, which is chosen as a conservative trade-off between accuracy and run-time. 

\subsection{Field Experiments}
\label{sec:experiments_field}

\begin{figure}
    \centering
    \includegraphics[width=0.475\textwidth]{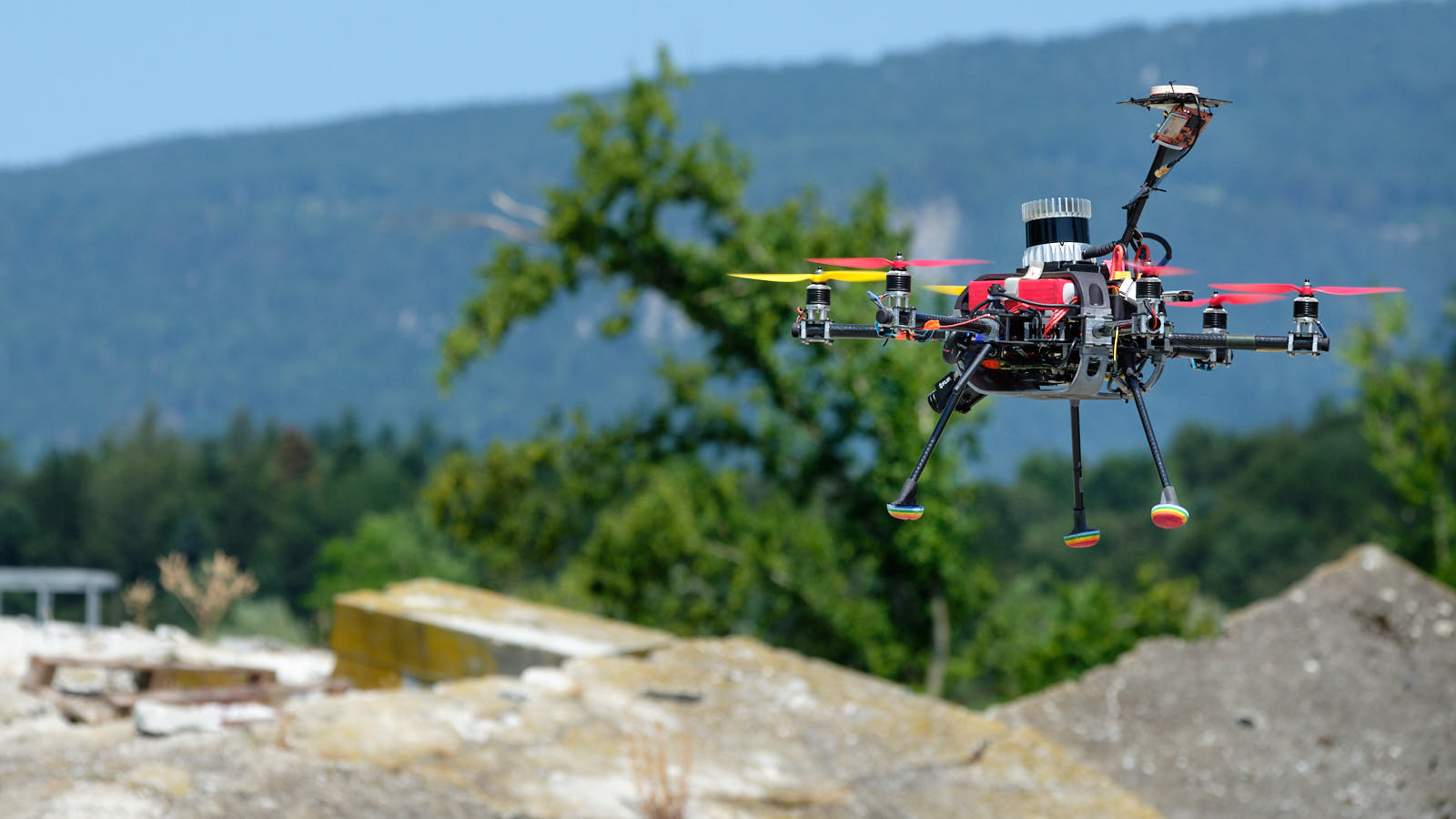}
    \caption{A hexacopter \ac{MAV} flying through a search and rescue training area (Sec.~\ref{sec:experiments_lidar}). The platform was equipped with a 64 beam lidar (visible), monocular camera, time-synchronized \ac{IMU}, and RTK-GNSS (used for trajectory evaluation only).}
    \label{fig:noliro_rubble}
    \vspace{-5mm}
\end{figure}

We present results collected on-board a hexacopter \ac{MAV} (shown in Fig.~\ref{fig:noliro_rubble}). All calculations are performed using an Intel NUC Core i7-8650U processor, which is carried by the platform. To demonstrate the flexibility of the proposed system we show results for both LiDAR and RGB-D-based sensor suites. LiDAR data is produced by an Ouster OS1 LiDAR. RGB-D data is produced by an Intel RealSense D415 Depth Camera. In all experiments a time-synchronized camera-imu setup is used with the visual-inertial odometry pipeline of Bl\"{o}sch et. al.~\cite{bloesch2017iterated} to provide odometry to the proposed system.

\subsubsection{Lidar-based Mapping}
\label{sec:experiments_lidar}

In this section the \ac{MAV} flew through a disaster area designed for training search and rescue personnel at Wangen an der Aare, Switzerland. Figure~\ref{fig:arche_reconstruction} shows an example reconstruction produced by \textit{voxgraph} on one of these datasets. We perform a quantitative evaluation of the proposed system performance. As gathering ground truth geometry of such a large space is difficult, we attached an RTK-GNSS system to collect accurate positioning information during the \ac{MAV}'s flight. We use this as ground truth against which we evaluate estimated trajectories. Note that the proposed system is map-centeric, in the sense the we do not compute an optimal trajectory. To obtain a trajectory for evaluation, we take the history of submap-relative odometry measurements~\refeq{eq:pose_history} and project them into the world frame using the optimal submap poses from~\refeq{eq:optimization}.

We compare the proposed system against state-of-the-art frameworks for trajectory estimation. In particular, we compare against LOAM~\cite{zhang2017low} (LiDAR) and VINS-MONO~\cite{qin2018vins} (visual-inertial). We also compare against ROVIO~\cite{bloesch2017iterated} (visual-inertial odometry), which provides odometry estimates to \textit{voxgraph}. This is included to show the performance of the proposed system \textit{without} submap registration.

\begin{figure}[t]
    \centering
    \begin{subfigure}[b]{0.45\columnwidth}
    \includegraphics[trim={0cm 0cm 0cm 0cm},clip,width=\columnwidth]{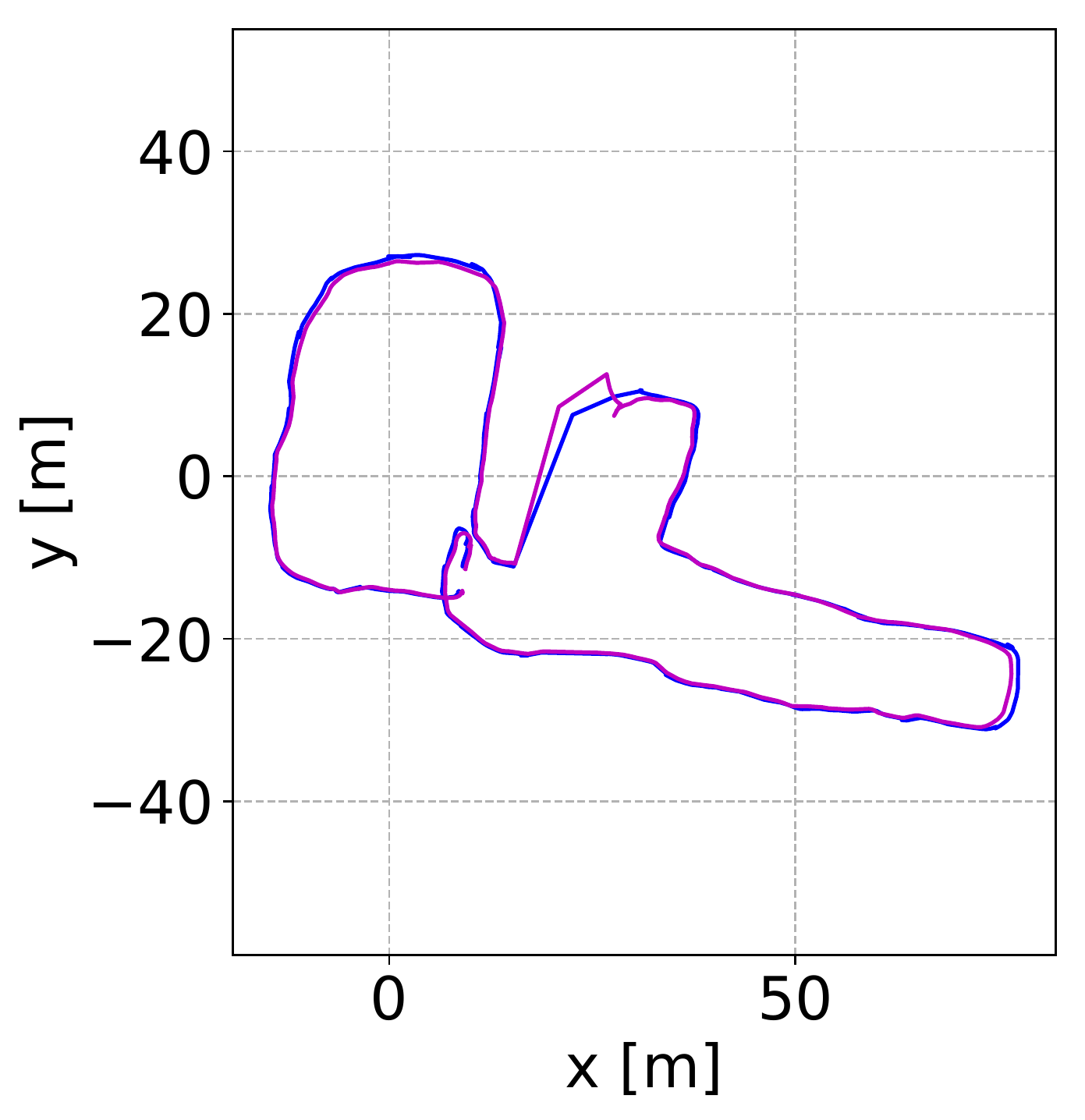}
    \end{subfigure}
    \begin{subfigure}[b]{0.45\columnwidth}
    \includegraphics[trim={0cm 0cm 0cm 0cm},clip,width=\columnwidth]{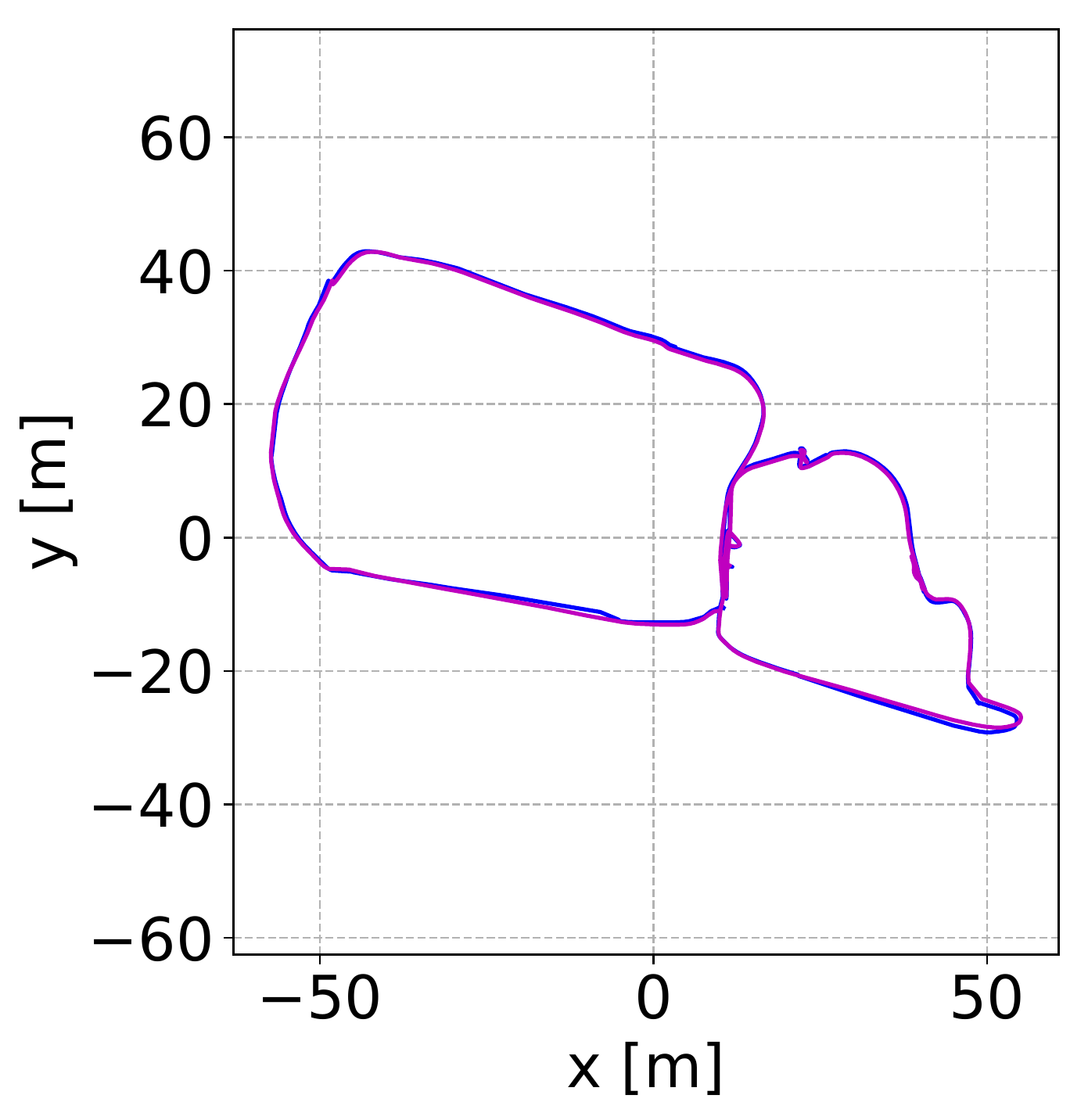}
    \end{subfigure}
    \begin{subfigure}[b]{0.45\columnwidth}
    \includegraphics[trim={0cm 0cm 0cm 0cm},clip,width=\columnwidth]{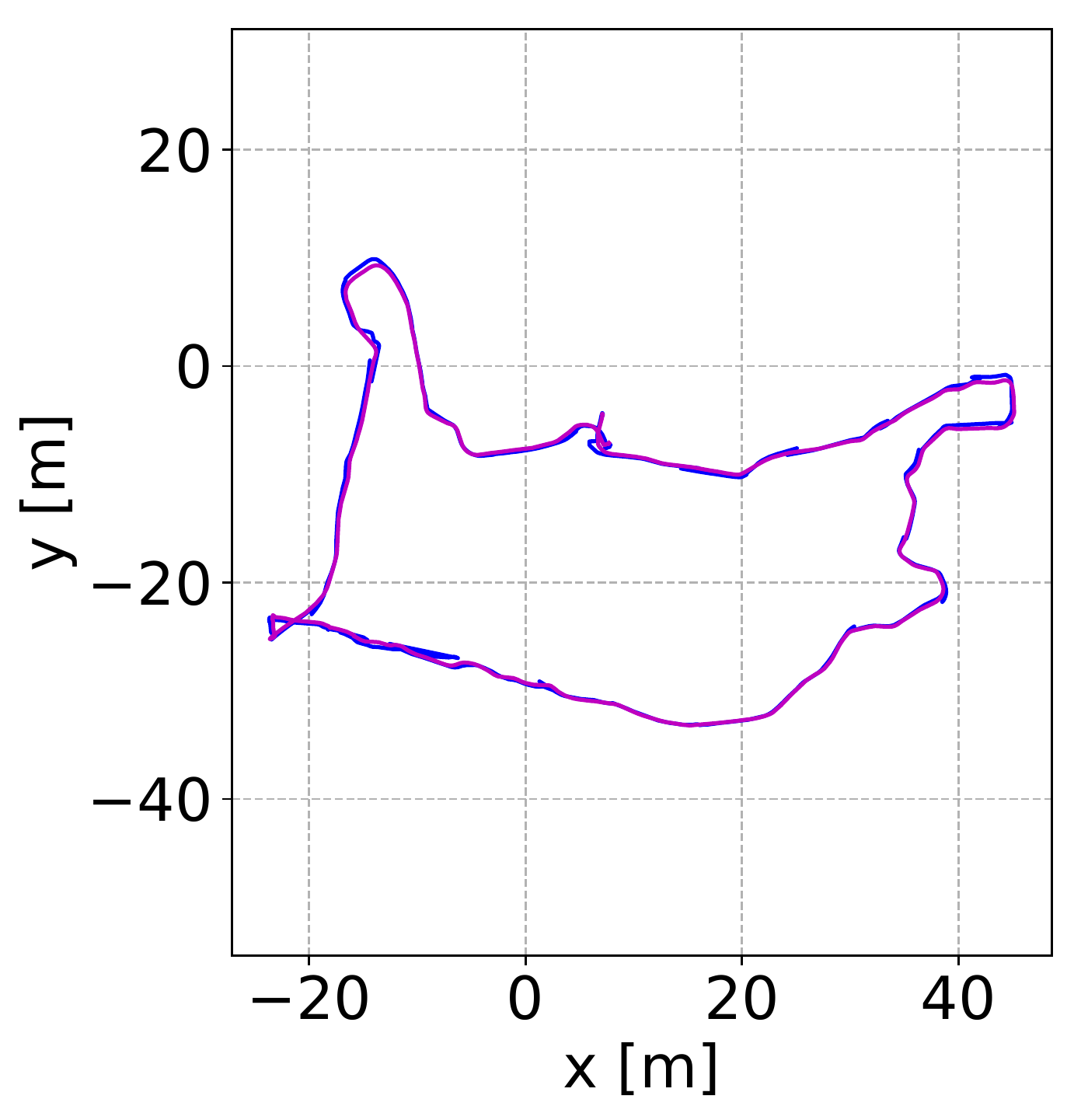}
    \end{subfigure}
    \begin{subfigure}[b]{0.45\columnwidth}
    \includegraphics[trim={0cm 0cm 0cm 0cm},clip,width=\columnwidth]{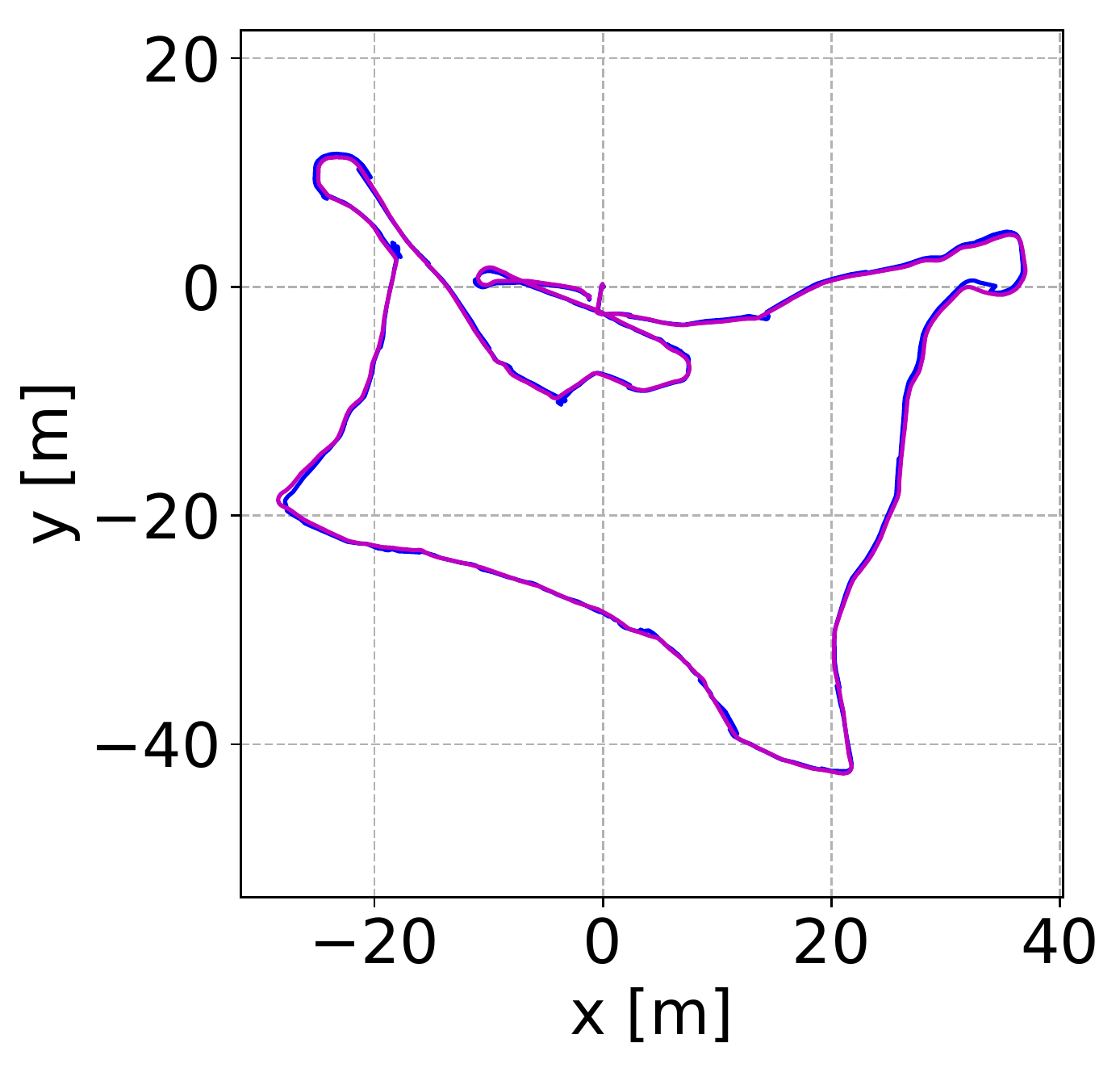}
    \end{subfigure}
    \caption{The trajectories from the field experiments in Sec.~\ref{sec:experiments_field} produced by an \ac{MAV} flying through a large outdoor space used to train search and rescue personnel. The plots show the \textit{voxgraph} estimated trajectory (blue), and RTK-GNSS measurements used to evaluate the system (pink).}
    \label{fig:trajectories}
    \vspace{-4mm}
\end{figure}

We evaluate the proposed and comparison frameworks using \ac{ATE}~\cite{zhang2018tutorial}. Prior to evaluation, we globally align the visual-inertial systems over 4 \ac{DOF} and LOAM over 6 \ac{DOF} (necessary because the estimated trajecory is not gravity aligned). The estimates produced by the proposed system, as well as the ground truth trajectories, are shown in Fig.~\ref{fig:trajectories}. We compute the \ac{RMSE} between corresponding samples on the estimated and ground truth trajectories. Both the proposed and comparison systems have random elements. We therefore run each trajectory 10 times per dataset and report average figures for both the \ac{RMSE} position error, as well as the CPU usage. Results of these evaluation are shown in Table~\ref{tab:arche}. The results show that the proposed system substantially outperforms the odometry estimated trajectory, as well as the comparison systems, on all datasets. \textit{Voxgraph} achieves 0.83\,\si{\meter}, 0.59\,\si{\meter}, 0.94\,\si{\meter}, 0.52\,\si{\meter} \ac{RMSE} drift on the four trajectories respectively.

The proposed system runs on $<3$~CPU threads on the processor carried by the \ac{MAV}. Note that 100\% in these terms corresponds to one (hyper) thread being fully utilized; the maximum computational load of the Intel NUC Core i7-8650U is therefore 800\%. While the proposed system uses more CPU than the frameworks we compared to, ours is the only one to compute a volumetric map, usable for planning and collision-checking, as part of the CPU usage. If we discount the time for computing the T- and ESDF, the actual optimization time is comparable or even less than the comparison systems.

\begin{figure}
    \centering
    \includegraphics[width=1.0\columnwidth]{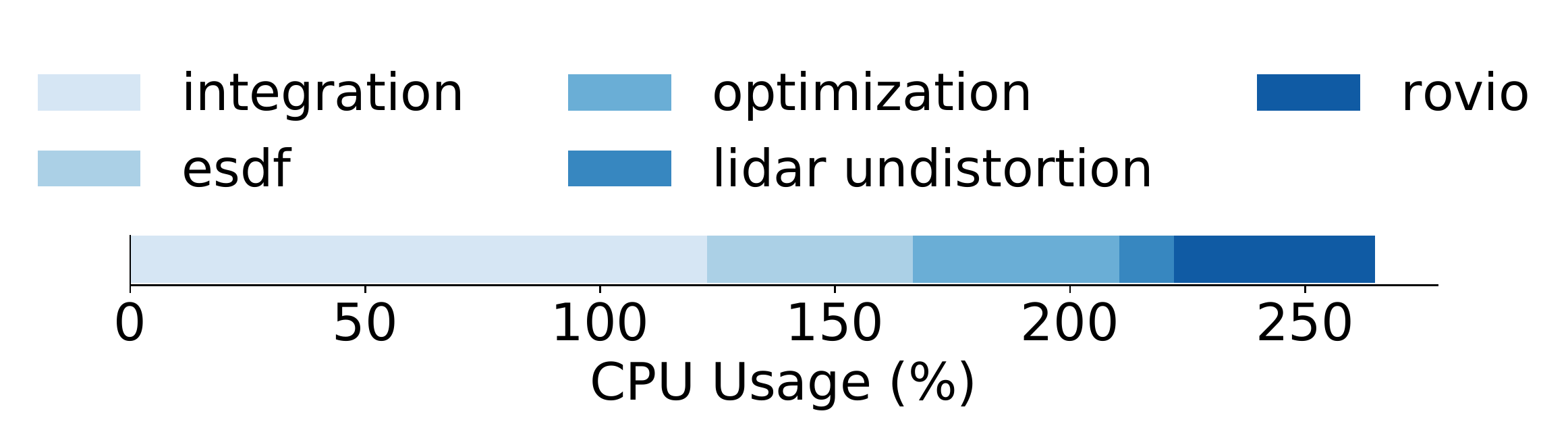}
    \caption{A breakdown of \textit{voxgraph} CPU usage during a typical \ac{MAV} flight from Sec.~\ref{sec:experiments_field}. The global optimization scheme suggested in this proposal consumes 44\% of a single CPU core.}
    \label{fig:cpu_breakdown}
    \vspace{-4mm}
\end{figure}
\begin{figure}[t]
    \centering
    \begin{subfigure}[b]{0.45\columnwidth}
    \includegraphics[trim={0cm 0cm 0cm 0cm},clip,width=\columnwidth]{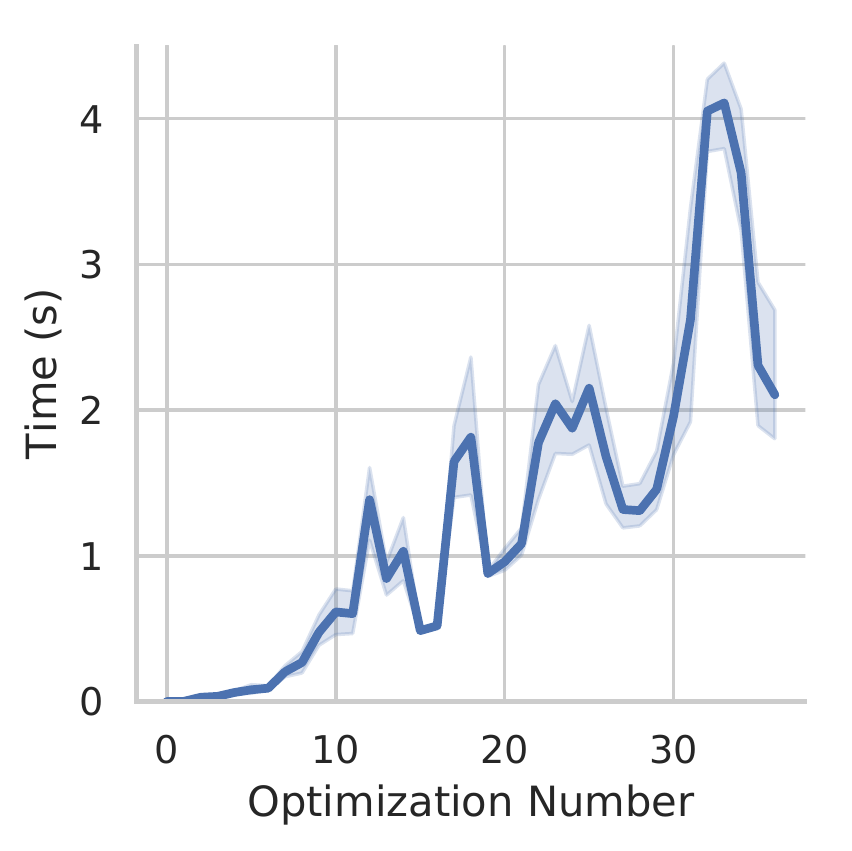}
    \end{subfigure}
    \begin{subfigure}[b]{0.45\columnwidth}
    \includegraphics[trim={0cm 0cm 0cm 0cm},clip,width=\columnwidth]{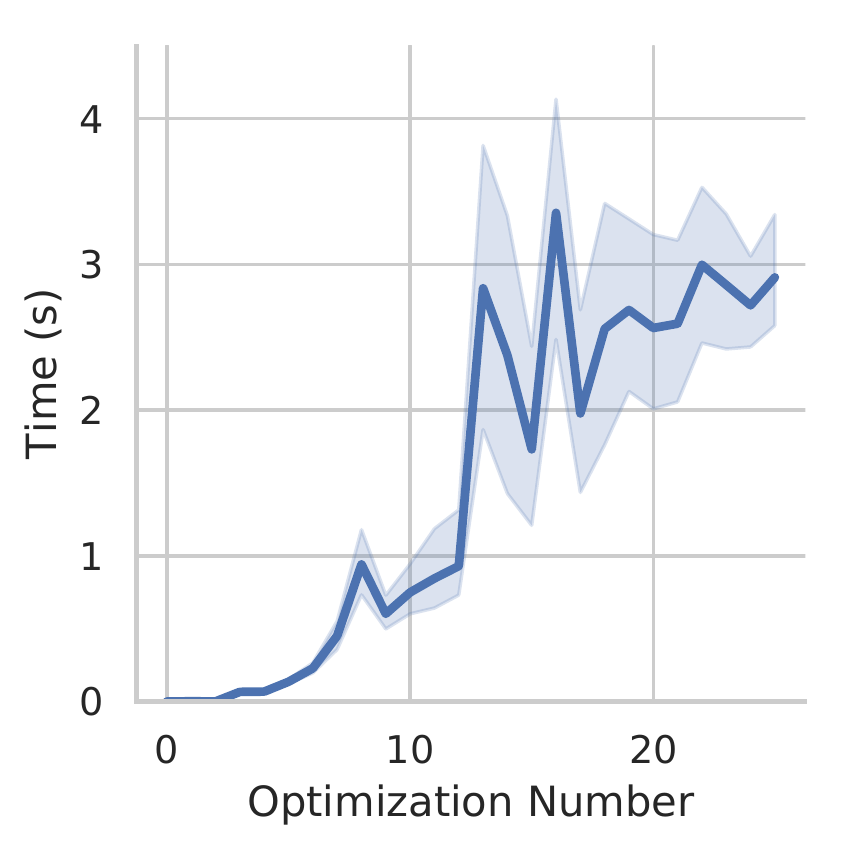}
    \end{subfigure}
    \begin{subfigure}[b]{0.45\columnwidth}
    \includegraphics[trim={0cm 0cm 0cm 0cm},clip,width=\columnwidth]{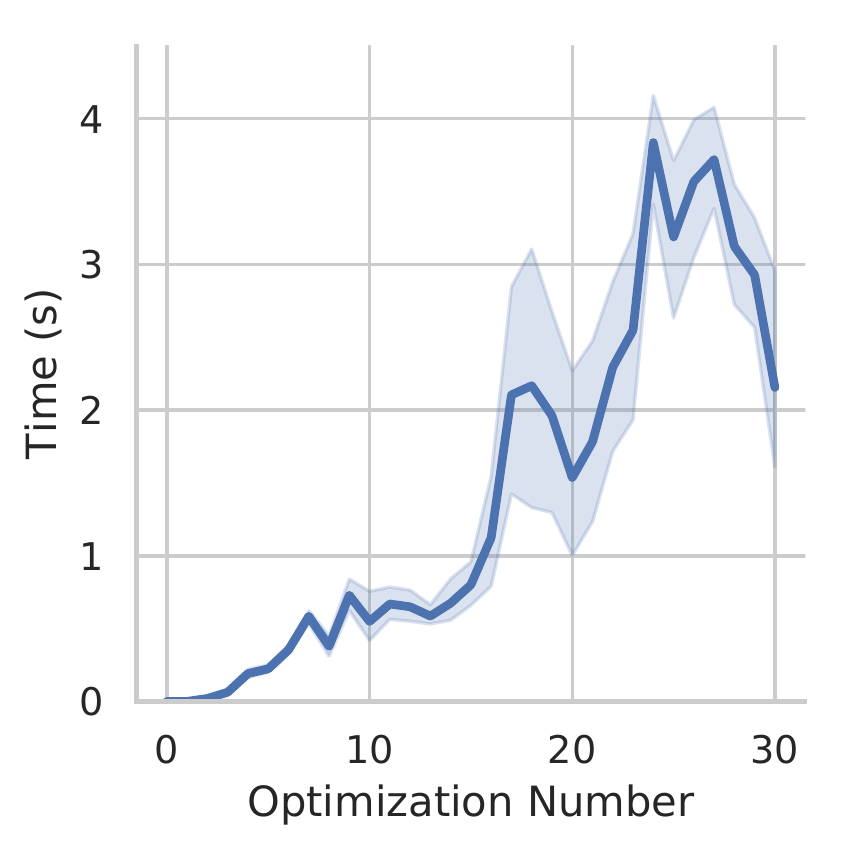}
    \end{subfigure}
    \begin{subfigure}[b]{0.45\columnwidth}
    \includegraphics[trim={0cm 0cm 0cm 0cm},clip,width=\columnwidth]{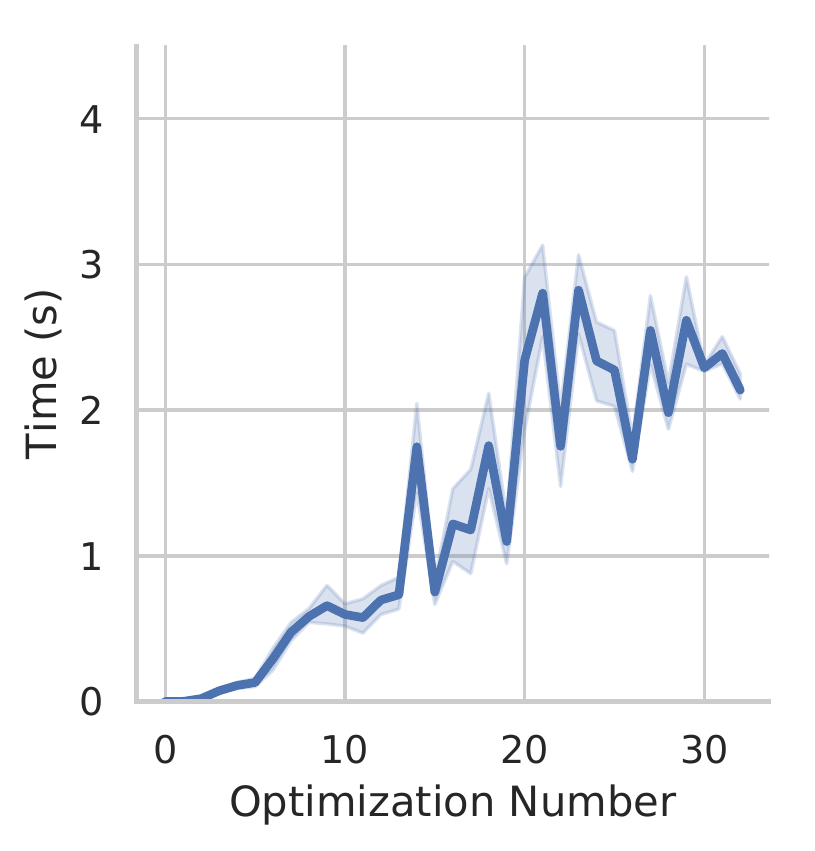}
    \end{subfigure}
    \caption{The global optimization times from the field experiments in Sec.~\ref{sec:experiments_field} produced by an \ac{MAV} flying through a large outdoor space used to train search and rescue personnel. Each data point is generated by a single global optimization, during one of 10 mapping trials of each of 4 trajectories.}
    \label{fig:optization_times}
    \vspace{-4mm}
\end{figure}

CPU usage for the proposed system is broken down into its major subroutines in Fig.~\ref{fig:cpu_breakdown}, namely pointcloud integration, \ac{ESDF} computation, pose-graph optimization, LiDAR undistortion, and ROVIO odometry. Global optimizations use 44\% of a single CPU core. Figure~\ref{fig:optization_times} shows global optimization run-times for each of the 4 trajectories, (again 10 trials per trajectory are used to generate the results). The maximum runtime for global optimization is ${\sim}4$\,\si{\second}.

\begin{table}[t]\vspace{0.5\intextsep}
\centering
\ra{1.3}
\begin{tabular}{@{}rrrrr|rrrr@{}}\toprule
\phantom{abc} & \multicolumn{4}{c}{RMSE (m)} & \multicolumn{4}{c}{CPU (\%)} \\
\cmidrule{2-9}
System & t0 & t1 & t2 & t3 & t0 & t1 & t2 & t3 \\
\midrule
ROVIO      & 4.55 & 1.30 & 3.52 & 2.25 & 47 & 47 & 48 & 47 \\
\textit{Voxgraph} & \textbf{0.83} & \textbf{0.59} & \textbf{0.94} & \textbf{0.52} & 265 & 230 & 305 & 286 \\
\midrule
Vins-Mono & 5.51 & - & 1.11 & - & 159 & - & 157 & - \\
Loam & 2.64 & 1.29 & 6.43 & 3.55 & 131 & 129 & 125 & 127 \\
\bottomrule
\end{tabular}
\caption{\ac{RMSE} for the field experiments conducted in Sec~\ref{sec:experiments_field}. The proposed method \textit{voxgraph} is compared against two state of the art systems for trajectory estimation, Vins-Mono~\cite{qin2018vins} and LOAM~\cite{zhang2017low}. The performance for ROVIO~\cite{bloesch2017iterated}, which provides the proposed system with odometry estimates, is also reported. This result can be interpreted as the front-end-only performance of the proposed system. Note the results denoted at $-$ indicate that the estimator diverged.}
\label{tab:arche}
\vspace{-4mm}
\end{table}

%

\subsubsection{RGB-D-based Mapping}
\label{sec:experiments_rgbd}

In order to validate how loop-closure constraints can be used in the proposed system, we applied our method to an indoor dataset~\cite{millane2018cblox}.
The dataset is collected by an \ac{MAV}, carrying an RGB-D camera and vi-sensor~\cite{nikolic2014synchronized}, flying through an underground industrial site at ETH Z{\"u}rich. The dataset contains significant odometry drift and multiple wide-baseline loops which cannot be corrected using submap registration alone. We therefore utilize appearance-based place recognition~\cite{GalvezTRO12} to generate loop-closure constraints from vi-sensor images. The dataset is processed on a desktop CPU with 5\,cm voxels. 

\begin{figure}[t]
    \centering
    \hspace*{-10mm}
    \begin{subfigure}[b]{0.2\textwidth}
        \includegraphics[width=\textwidth]{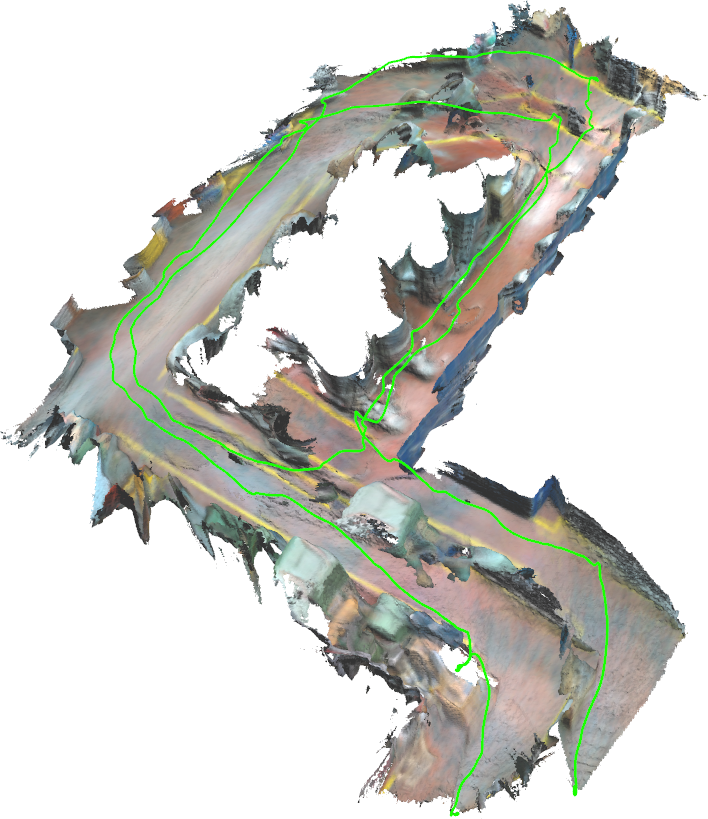}
        \vspace{0.1mm}
    \end{subfigure}
    \begin{subfigure}[b]{0.2\textwidth}
        \includegraphics[width=\textwidth]{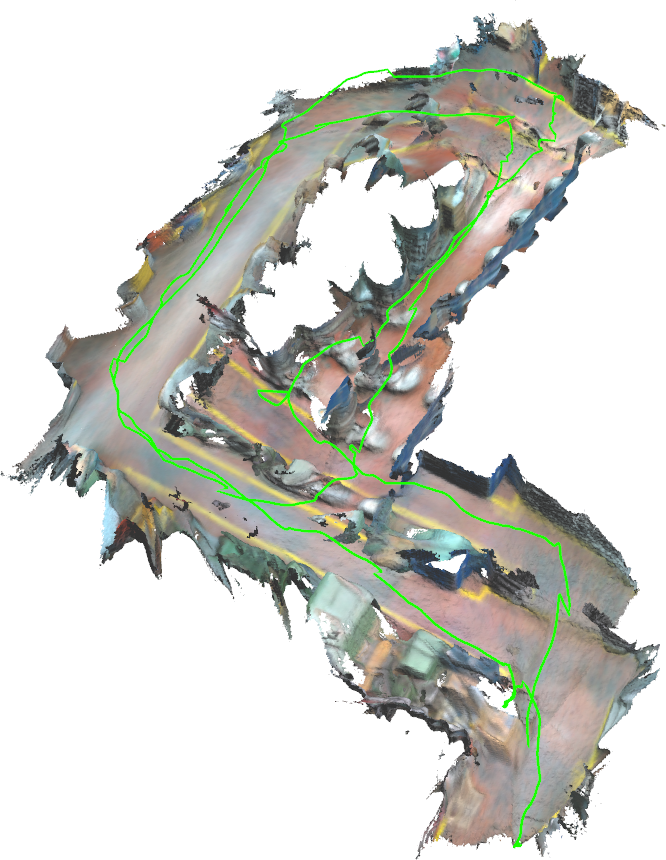}
    \end{subfigure}
    \hspace*{-10mm}
    \begin{subfigure}[b]{0.2\textwidth}
        \includegraphics[width=\textwidth]{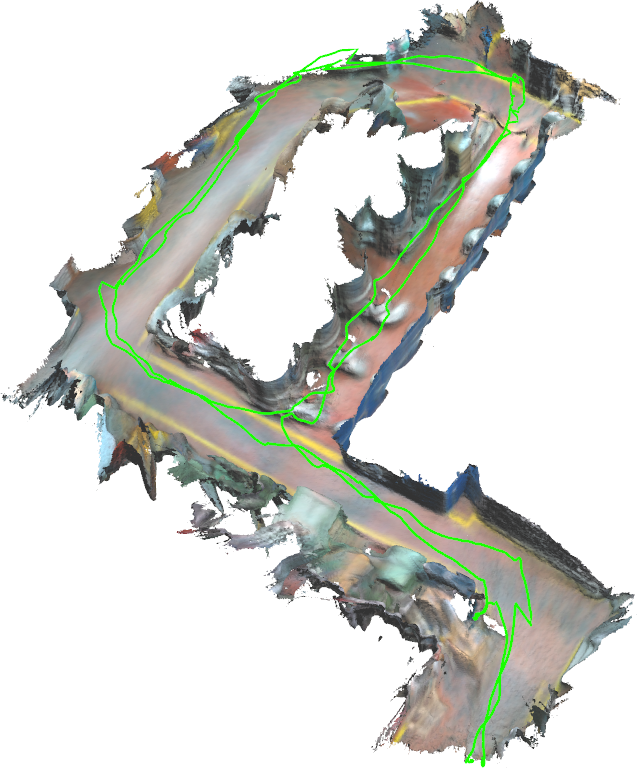}
    \end{subfigure}
    \caption{Reconstruction of an industrial environment using data collected from two flights of an \ac{MAV} using the proposed approach and RGB-D camera data. Consistency is maintained in the underlying representation through alignment of \ac{SDF} subvolumes, even though the input odometry drifts over time. The top-left figure is the reconstruction using visual-inertial odometry-only, the top right is with registration constraints, and the bottom figure is using registration \textit{and} loop-closure constraints.}
    \label{fig:machine_hall}
    \vspace{-5mm}
\end{figure}

Figure~\ref{fig:machine_hall} shows the results of mapping using the visual-inertial odometry system~\cite{bloesch2017iterated} alone, and \textit{voxgraph} with and  without loop closures. In the first two configurations the generated map shows significant distortion, resulting from the \ac{MAV} revisiting a location with an inconsistent pose estimate. Inclusion of submap-relative loop-closure constraints (Sec.~\ref{sec:backend_loopclosure}), generates corrects gross map distortions.


%% file: sections/conclusion.tex
\section{Conclusion}
\label{sec:conclusion}

In this paper we presented \textit{voxgraph}, a novel framework for \ac{SDF}-based reconstruction, aimed at producing globally consistent maps. The system represents the observed scene as a collection of overlapping \ac{SDF} submaps. The system is a map-centric approach in which global consistency is maintained through geometric alignment of submaps. For each submap we compute a refined distance field, the \ac{ESDF}, as well as a set of zero-level iso-surface points in the front-end. This information is used to generate efficient, correspondence-free registration constraints between submap pairs in the back-end. We propose to include only a random sub-sample of residual terms during optimization to dramatically increase computational efficiency. The efficiency of our registration constraints allow us to perform global optimization regularly, for instance after the completion of each additional submap. 

The result of the proposed ideas is that the system creates consistent, large-scale maps in real-time on a lightweight computer. We show in a simulation study that constraint sub-sampling has the desired effect of boosting computational efficiency, without dramatically impacting reconstruction accuracy. In field experiments, we show that despite not explicitly estimating an optimal trajectory, and spending the majority of our computation on generating a global volumetric map, the proposed system outperforms a state-of-the-art visual \ac{SLAM} system and a top-performing LiDAR system, even when evaluated on trajectory error alone. Furthermore, \textit{Voxgraph} is released as open-source.


%% file: main.bbl
\begin{thebibliography}{10}
\providecommand{\url}[1]{#1}
\csname url@samestyle\endcsname
\providecommand{\newblock}{\relax}
\providecommand{\bibinfo}[2]{#2}
\providecommand{\BIBentrySTDinterwordspacing}{\spaceskip=0pt\relax}
\providecommand{\BIBentryALTinterwordstretchfactor}{4}
\providecommand{\BIBentryALTinterwordspacing}{\spaceskip=\fontdimen2\font plus
\BIBentryALTinterwordstretchfactor\fontdimen3\font minus
  \fontdimen4\font\relax}
\providecommand{\BIBforeignlanguage}[2]{{%
\expandafter\ifx\csname l@#1\endcsname\relax
\typeout{** WARNING: IEEEtran.bst: No hyphenation pattern has been}%
\typeout{** loaded for the language `#1'. Using the pattern for}%
\typeout{** the default language instead.}%
\else
\language=\csname l@#1\endcsname
\fi
#2}}
\providecommand{\BIBdecl}{\relax}
\BIBdecl

\bibitem{cadena2016past}
C.~Cadena, L.~Carlone, H.~Carrillo, Y.~Latif, D.~Scaramuzza, J.~Neira, I.~Reid,
  and J.~J. Leonard, ``Past, present, and future of simultaneous localization
  and mapping: Toward the robust-perception age,'' \emph{T-RO}, vol.~32, no.~6,
  pp. 1309--1332, 2016.

\bibitem{mur2015orb}
R.~Mur-Artal, J.~M.~M. Montiel, and J.~D. Tardos, ``Orb-slam: a versatile and
  accurate monocular slam system,'' \emph{T-RO}, vol.~31, no.~5, pp.
  1147--1163, 2015.

\bibitem{schneider2018maplab}
T.~Schneider, M.~T. Dymczyk, M.~Fehr, K.~Egger, S.~Lynen, I.~Gilitschenski, and
  R.~Siegwart, ``maplab: An open framework for research in visual-inertial
  mapping and localization,'' \emph{RAL}, 2018.

\bibitem{burri2015real}
M.~Burri, H.~Oleynikova, M.~W. Achtelik, and R.~Siegwart, ``Real-time
  visual-inertial mapping, re-localization and planning onboard mavs in unknown
  environments,'' in \emph{IROS}.\hskip 1em plus 0.5em minus 0.4em\relax IEEE,
  2015, pp. 1872--1878.

\bibitem{qin2018vins}
T.~Qin, P.~Li, and S.~Shen, ``Vins-mono: A robust and versatile monocular
  visual-inertial state estimator,'' \emph{T-RO}, vol.~34, no.~4, pp.
  1004--1020, 2018.

\bibitem{engel2014lsd}
J.~Engel, T.~Sch{\"o}ps, and D.~Cremers, ``Lsd-slam: Large-scale direct
  monocular slam,'' in \emph{ECCV}.\hskip 1em plus 0.5em minus 0.4em\relax
  Springer, 2014, pp. 834--849.

\bibitem{whelan2012kintinuous}
T.~Whelan, M.~Kaess, M.~Fallon, H.~Johannsson, J.~J. Leonard, and J.~McDonald,
  ``Kintinuous: Spatially extended kinectfusion,'' in \emph{{AAAI} 2012}, 2012.

\bibitem{whelan2015elasticfusion}
T.~Whelan, S.~Leutenegger, R.~Salas-Moreno, B.~Glocker, and A.~Davison,
  ``Elasticfusion: Dense slam without a pose graph,'' in \emph{RSS}.\hskip 1em
  plus 0.5em minus 0.4em\relax Robotics: Science and Systems, 2015.

\bibitem{hornung2013octomap}
A.~Hornung, K.~M. Wurm, M.~Bennewitz, C.~Stachniss, and W.~Burgard, ``Octomap:
  An efficient probabilistic 3d mapping framework based on octrees,''
  \emph{Autonomous robots}, vol.~34, no.~3, pp. 189--206, 2013.

\bibitem{curless1996volumetric}
B.~Curless and M.~Levoy, ``\BIBforeignlanguage{en}{A volumetric method for
  building complex models from range images},'' in
  \emph{\BIBforeignlanguage{en}{SIGGRAPH}}.\hskip 1em plus 0.5em minus
  0.4em\relax ACM Press, 1996, pp. 303--312.

\bibitem{izadi2011kinectfusion}
S.~Izadi, D.~Kim, O.~Hilliges, D.~Molyneaux, R.~Newcombe, P.~Kohli, J.~Shotton,
  S.~Hodges, D.~Freeman, A.~Davison \emph{et~al.}, ``Kinectfusion: real-time 3d
  reconstruction and interaction using a moving depth camera,'' in
  \emph{UIST}.\hskip 1em plus 0.5em minus 0.4em\relax ACM, 2011, pp. 559--568.

\bibitem{lin2018autonomous}
Y.~Lin, F.~Gao, T.~Qin, W.~Gao, T.~Liu, W.~Wu, Z.~Yang, and S.~Shen,
  ``Autonomous aerial navigation using monocular visual-inertial fusion,''
  \emph{JFR}, vol.~35, no.~1, pp. 23--51, 2018.

\bibitem{oleynikova2017voxblox}
H.~Oleynikova, Z.~Taylor, M.~Fehr, R.~Siegwart, and J.~Nieto, ``Voxblox:
  Incremental 3d euclidean signed distance fields for on-board mav planning,''
  in \emph{IROS}.\hskip 1em plus 0.5em minus 0.4em\relax IEEE, 2017, pp.
  1366--1373.

\bibitem{wagner2014graph}
R.~Wagner, U.~Frese, and B.~B{\"a}uml, ``Graph slam with signed distance
  function maps on a humanoid robot,'' in \emph{IROS}.\hskip 1em plus 0.5em
  minus 0.4em\relax IEEE, 2014, pp. 2691--2698.

\bibitem{oleynikova2018safe}
H.~Oleynikova, Z.~Taylor, R.~Siegwart, and J.~Nieto, ``Safe local exploration
  for replanning in cluttered unknown environments for microaerial vehicles,''
  \emph{RAL}, vol.~3, no.~3, pp. 1474--1481, 2018.

\bibitem{ratliff2009chomp}
N.~Ratliff, M.~Zucker, J.~A. Bagnell, and S.~Srinivasa, ``Chomp: Gradient
  optimization techniques for efficient motion planning,'' in
  \emph{ICRA}.\hskip 1em plus 0.5em minus 0.4em\relax figshare, 2009.

\bibitem{dai2017bundlefusion}
A.~Dai, M.~Nie{\ss}ner, M.~Zollh{\"o}fer, S.~Izadi, and C.~Theobalt,
  ``Bundlefusion: Real-time globally consistent 3d reconstruction using
  on-the-fly surface reintegration,'' \emph{ACM ToG}, vol.~36, no.~3, p.~24,
  2017.

\bibitem{maier2017efficient}
R.~Maier, R.~Schaller, and D.~Cremers, ``Efficient online surface correction
  for real-time large-scale 3d reconstruction,'' \emph{BMVC}, 2017.

\bibitem{han2018flashfusion}
H.~Lei and L.~Fang, ``Flashfusion: Real-time globally consistent dense 3d
  reconstruction using cpu computing,'' in \emph{Robotics: {Science} and
  {Systems}}.\hskip 1em plus 0.5em minus 0.4em\relax RSS, 2018.

\bibitem{henry2013patch}
P.~Henry, D.~Fox, A.~Bhowmik, and R.~Mongia, ``Patch volumes:
  Segmentation-based consistent mapping with rgb-d cameras,'' in
  \emph{3DV}.\hskip 1em plus 0.5em minus 0.4em\relax IEEE, 2013, pp. 398--405.

\bibitem{kahler2016real}
O.~K{\"a}hler, V.~A. Prisacariu, and D.~W. Murray, ``Real-time large-scale
  dense 3d reconstruction with loop closure,'' in \emph{ECCV}.\hskip 1em plus
  0.5em minus 0.4em\relax Springer, 2016, pp. 500--516.

\bibitem{choi2015robust}
S.~Choi, Q.-Y. Zhou, and V.~Koltun, ``Robust reconstruction of indoor scenes,''
  in \emph{CVPR}, 2015, pp. 5556--5565.

\bibitem{hess2016real}
W.~Hess, D.~Kohler, H.~Rapp, and D.~Andor, ``Real-time loop closure in 2d lidar
  slam,'' in \emph{ICRA}.\hskip 1em plus 0.5em minus 0.4em\relax IEEE, 2016,
  pp. 1271--1278.

\bibitem{fioraio2015large}
N.~Fioraio, J.~Taylor, A.~Fitzgibbon, L.~Di~Stefano, and S.~Izadi,
  ``Large-scale and drift-free surface reconstruction using online subvolume
  registration,'' in \emph{CVPR}, 2015, pp. 4475--4483.

\bibitem{niessner2013real}
M.~Nie{\ss}ner, M.~Zollh{\"o}fer, S.~Izadi, and M.~Stamminger, ``Real-time 3d
  reconstruction at scale using voxel hashing,'' \emph{ACM ToG}, vol.~32,
  no.~6, p. 169, 2013.

\bibitem{millane2018cblox}
A.~Millane, Z.~Taylor, H.~Oleynikova, J.~Nieto, R.~Siegwart, and C.~Cadena,
  ``C-blox: A scalable and consistent tsdf-based dense mapping approach,'' in
  \emph{IROS}, 2018.

\bibitem{lorensen_1987_marching-cubes}
W.~E. Lorensen and H.~E. Cline, ``Marching cubes: {A} high resolution 3d
  surface construction algorithm,'' \emph{SIGGRAPH}, 1987.

\bibitem{GalvezTRO12}
D.~G\'alvez-L\'opez and J.~D. Tard\'os, ``Bags of binary words for fast place
  recognition in image sequences,'' \emph{R-RO}, vol.~28, no.~5, pp.
  1188--1197, October 2012.

\bibitem{christer_ericson_2005_collision-detection}
{Christer Ericson}, \emph{Real-time collision detection}, ser. Morgan
  {Kaufmann} series in interactive 3D technology.\hskip 1em plus 0.5em minus
  0.4em\relax Amsterdam; Boston: Elsevier, 2005.

\bibitem{dellaert2017factor}
F.~Dellaert, M.~Kaess \emph{et~al.}, ``Factor graphs for robot perception,''
  \emph{Foundations and Trends{\textregistered} in Robotics}, vol.~6, no. 1-2,
  pp. 1--139, 2017.

\bibitem{kang_2006_color-technology}
H.~R. Kang, \emph{Computational color technology}.\hskip 1em plus 0.5em minus
  0.4em\relax Bellingham, Wash: SPIE Press, 2006.

\bibitem{olson_2006_fast-iterative}
E.~Olson, J.~Leonard, and S.~Teller, ``Fast iterative alignment of pose graphs
  with poor initial estimates,'' in \emph{ICRA}, 2006.

\bibitem{bloesch2017iterated}
M.~Bloesch, M.~Burri, S.~Omari, M.~Hutter, and R.~Siegwart, ``Iterated extended
  kalman filter based visual-inertial odometry using direct photometric
  feedback,'' \emph{IJRR}, vol.~36, no.~10, pp. 1053--1072, 2017.

\bibitem{zhang2017low}
J.~Zhang and S.~Singh, ``Low-drift and real-time lidar odometry and mapping,''
  \emph{Autonomous Robots}, vol.~41, no.~2, pp. 401--416, 2017.

\bibitem{zhang2018tutorial}
Z.~Zhang and D.~Scaramuzza, ``A tutorial on quantitative trajectory evaluation
  for visual (-inertial) odometry,'' in \emph{IROS}.\hskip 1em plus 0.5em minus
  0.4em\relax IEEE, 2018, pp. 7244--7251.

\bibitem{nikolic2014synchronized}
J.~Nikolic, J.~Rehder, M.~Burri, P.~Gohl, S.~Leutenegger, P.~T. Furgale, and
  R.~Siegwart, ``A synchronized visual-inertial sensor system with fpga
  pre-processing for accurate real-time slam,'' in \emph{ICRA}.\hskip 1em plus
  0.5em minus 0.4em\relax IEEE, 2014, pp. 431--437.

\end{thebibliography}
